\documentclass{iopjournal}
\usepackage{times}
\usepackage{graphicx} 
\usepackage{amsmath} 
\usepackage{algorithm}
\usepackage{algorithmic}
\usepackage{booktabs}
\usepackage[utf8]{inputenc}

\begin{document}

\title{Prediction of Runtime Parameters of Parallel Chemistry Applications via Active and Generative Learning}

\author{Tanzila Tabassum$^1$, Omer Subasi$^2$, Ajay Panyala$^2$, Epiya Ebiapia$^1$, Gerald Baumgartner$^1$, Erdal Mutlu$^2$, P (Saday) Sadayappan$^3$ and Karol Kowalski$^2$}

\affil{$^1$Louisiana State University, Baton Rouge, Louisiana, USA\\}
\affil{$^2$Pacific Northwest National Laboratory, Richland, Washington, USA\\}
\affil{$^3$University of Utah, Salt Lake City, Utah, USA}

\email{ttabas1@lsu.edu, omer.subasi@pnnl.gov, ajay.panyala@pnnl.gov, eebiap1@lsu.edu, gb@lsu.edu, erdal.mutlu@pnnl.gov, saday@cs.utah.edu, Karol.Kowalski@pnnl.gov}


\begin{abstract}
    In this work, we develop two main Machine Learning 
    based approaches to predict the runtime parameters of
    highly scalable parallel chemistry computations.
    These approaches employ active and generative learning together with the empirically determined gradient boosted regression tree models chosen among a rich suite of machine learning models. 
    
    When evaluated on Coupled-Cluster with Singles and Doubles computations, our models achieve a mean absolute error percentage (MAPE) as low as 0.023 and a coefficient of determination as high as 99.9\%. Furthermore, when combined with active learning to mitigate the lack of large amounts of training data, our models score a MAPE about 0.2 with 20--25\% of the original dataset.
    
\end{abstract}
\section{Introduction}
Supercomputers enable large-scale electronic-structure calculations using computational chemistry software. These applications support massively parallel execution on distributed-memory supercomputing systems for high performance with large node counts. Execution time for massively parallel computational chemistry runs depends on the problem size, the target machine, and runtime settings. In practice, users must choose these parameters prior to submitting jobs. However, identifying optimal runtime settings through trial runs is impractical because suboptimal configurations can lead to wasting significant node-hours under limited allocation. Accurate runtime prediction using machine learning can help users with configuration selection and managing budgets. In this work, we use runtime measurements collected from executions of Coupled Cluster with Singles and Doubles (CCSD) on premier supercomputing systems.
We develop and evaluate ML-based runtime prediction models to answer three user inquiries: (i) Shortest-Time Question (STQ): When sufficient historical runtime measurements for a target application and machine are available, which node count and tile size minimize execution time  for a given problem size? (ii) When only limited runtime data are available for running a target application on a target supercomputer, which combinations of problem sizes, node counts, and tile sizes should be used for test runs to achieve the most accurate runtime prediction? (iii) Budget Question (BQ): Given a budget (CPU/GPU hours, node-hours, or energy), which runtime configurations adhere to the budget?

We consider two scenarios based on the availability of data for CCSD users across three premier Department of Energy (DOE) supercomputing platforms: Aurora (ALCF)~\cite{alcf_aurora}, Frontier (OLCF)~\cite{olcf_frontier}, and Perlmutter (NERSC)~\cite{nersc_pm}. In the first scenario, historical measurements are available on a target platform across problem sizes and runtime settings such as node count and tile size.
With sufficient data available, we train several supervised machine learning models and evaluate their comparative accuracy via cross-validation. We evaluate the models using mean absolute error (MAE), mean absolute percentage error (MAPE), and the coefficient of determination (R²) and find that Gradient Boosting regression provides the most accurate runtime predictions for both STQ and BQ goals. For example, on ALCF Aurora~\cite{alcf_aurora} our Gradient Boosting model achieves an STQ MAPE of 0.023 with MAE 2.36 and $R^2$ = 0.999, and for BQ, it achieves an MAPE of 0.12 with MAE 0.41 and $R^2$ = 0.979.
In the second scenario, historical measurements are limited, and collecting additional data from supercomputer runs is expensive.
To achieve accurate runtime prediction models even with a small number of measurements, we use two approaches. First, we apply active learning to select informative configurations for test runs. The query strategies for active learning that we implement are uncertainty sampling and query by committee. Second, we utilize generative models, Gaussian Copula and CTGAN, to augment the training dataset with synthetic data points.
Using active learning, we achieve STQ MAPE around 0.2 with roughly 20-25\% of the original dataset.

The main contributions of our study include:
\begin{itemize}
    \item Development and implementation of a suite of high-performant predictive ML models for runtime parameter prediction for the CCSD application,
    \item The use of active and generative learning to mitigate the issue of insufficient number of training executions, and 
    \item A thorough evaluation of the ML models based on performance data collected from real CCSD calculations on three premier DOE supercomputers: Aurora (ALCF), Frontier (OLCF), and Perlmutter (NERSC).
\end{itemize}

Our manuscript is organized as follows:
Section \ref{background_and_literature} provides background
and related work.
Section \ref{our_approach} presents our proposed approach.
Section \ref{experimental_evaluation} details the experimental
evaluation of our approaches and the results.
Finally, Section \ref{conclusion} concludes our study.

\section{Background and related work}
\label{background_and_literature}

Correlated electronic-structure methods, particularly Coupled Cluster with Singles and Doubles (CCSD)~\cite{ccsd1,ccsd2,ccsd3}, put significant computational demands on supercomputers. For a fixed basis set, the computational cost scales as \(N^6\)~\cite{ccsd1}.
 The execution cost of CCSD is dominated by the tensor operations and the communication among the distributed processes.
Therefore, previous works have focused on how to express, generate, and optimize tensor contractions for many-body methods. The Tensor Contraction Engine (TCE) introduced automated generation of tensor-contraction code for coupled-cluster and related methods~\cite{hirata2003tce,auer2006tce}. Similarly, related work~\cite{baumgartner2005synthesis} studied synthesizing high-performance parallel programs for ab initio quantum chemistry models expressed through tensor operations. 
Earlier work on tensor contraction optimization showed that the cost of many-body tensor expressions depends on data layout transformation, tensor permutation, and library-call selection~\cite{hartono2009tensor,lu2012layout}. More recently, TAMM has provided a tensor framework for expressing and executing many-body methods on distributed and heterogeneous systems that separates tensor-expression specification from execution~\cite{tamm_jcp_2023}. 

These developments, combined with greater access to computational resources, enable simulations of increasingly large and complex molecular systems, but also make the selection and allocation of computing resources more challenging. Resource estimation in quantum chemistry has traditionally been guided by theoretical scaling laws and empirical heuristics. Those are useful for understanding overall complexity, but they are often not sufficient for predicting actual runtime or optimal configurations such as block size, as well as hardware parameters, such as node count and GPUs. This is because of communication overhead, load imbalance, and effects of GPU on modern heterogeneous platforms that make manual tuning challenging. 
This is why parameter selection in HPC has often been treated as an autotuning problem. OpenTuner finds optimal parameters through an ensemble of search techniques~\cite{ansel2014opentuner}, while GPTune uses multitask learning to identify high-performing configurations for expensive exascale applications~\cite{liu2021gptune}.

Machine learning has been used in HPC for application performance prediction~\cite{yokelson2023hpp} and power modeling~\cite{bugbee2017power}, and in computational chemistry for computational cost estimation, including wall-time prediction for single-point, geometry-optimization, and transition-state calculations~\cite{heinen2020mlcostqc} and computational-time forecasting for DFT/TDDFT calculations~\cite{ma2021dfttime}. However, building accurate predictive models requires enough historical data obtained with expensive supercomputer runs. To tackle scarcity of data, we have explored active learning and synthetic data generation in our work. The active learning method selects which samples to label next so that fewer samples are needed for training predictive models. Uncertainty sampling and query-by-committee are two such strategies~\cite{settles2009active,seung1992query}. Prior work used active learning in HPC to reduce the number of expensive evaluations needed for surrogate-based performance tuning~\cite{balaprakash2013al}.

In our work, we use active learning to select informative CCSD configurations for runtime prediction. 
Generative learning, in contrast, aims to augment the training set by synthesizing plausible data points using models like Gaussian copulas or GANs~\cite{SDV,CTGAN2019}. These methods have been used in domains such as materials science~\cite{nagatani2022data}, fluid dynamics~\cite{kim2020unsupervised,feng2021data}, and molecular simulations~\cite{noe2020machine} to overcome data scarcity.
In this paper, we combine these ideas to develop an ML-based framework for answering user-facing questions about runtime and computational cost before a simulation is launched. The framework incorporates a range of regression models, along with active and generative strategies, to handle different levels of data availability. We evaluate the approach using data collected from real CCSD calculations on three premier Department of Energy (DOE) supercomputing platforms: Aurora (ALCF)~\cite{alcf_aurora}, Frontier (OLCF)~\cite{olcf_frontier}, and Perlmutter (NERSC)~\cite{nersc_pm}.

\section{Our proposed approaches}
\label{our_approach}

Predicting CCSD runtimes requires both an accurate surrogate model and a strategy for building that model when only limited measurements are available. We, therefore, consider a set of regression models, tune them for the target datasets, and use the best-performing models for the baseline experiments using the entire dataset, as well as the active-learning and generative-learning studies. The resulting models are used to answer two practical questions: which configuration gives the shortest runtime, and which configurations satisfy a given computational budget? Finally, we analyze the mean, variance, and standard deviation of runtimes in each selected or generated batch to examine whether the batches are dominated by short-running configurations, long-running configurations, or a mixture of both.

\subsection{Employed ML models}
\subsubsection*{Polynomial Regression (PR)} is used when the relationship between an independent variable \(x\) and a dependent variable \(y\) is nonlinear. It is an extension of multiple linear regression because, despite the nonlinear terms in \(x\), the model is linear in its coefficients. The complexity of polynomial regression is determined by the degree of the polynomial. Increasing the degree allows the model to fit more complex data; however, it also becomes more sensitive to noise and can lead to overfitting. In addition, the number of polynomial terms can grow quickly with an increase of the number of input variables.

\subsubsection*{Kernel Ridge Regression (KRR)} extends ridge regression by applying the kernel trick to fit a linear model in a high-dimensional feature space defined by the kernel. Thus, KRR learns nonlinear relationships in the original data while keeping the same \(\ell_{2}\) (ridge) regularization to prevent overfitting. The advantage of KRR is that it can model nonlinear relationships without explicit computations of the transformed high-dimensional feature space. KRR is fast because it uses a closed-form mathematical equation rather than iterative training. However, for large datasets, KRR introduces high computational cost of $O(n^3)$, and memory requirement of $O(n^2)$.
The performance of KRR is determined by the kernel function and the regularization parameter.

\subsubsection*{Decision Trees (DTs)} operate by partitioning the feature space into a set of rectangular regions based on rules applied to individual features. For regression tasks, a DT predicts the average target value of the training instances in each leaf node. This average value of a particular leaf node is then assigned as the prediction for new instances that are sorted into that same node. The tree is constructed recursively by selecting splits that reduce prediction error at each step. This recursive splitting continues until a stopping condition is reached, such as a maximum tree depth or a minimum number of samples in a node. DTs are highly interpretable and capable of modeling complex feature interaction without the need for data normalization. However, they are prone to instability due to minor changes in the data.

\subsubsection*{Random Forests (RFs)} are an ensemble technique that uses bootstrapping method~\cite{breiman2001random} to improve generalization by constructing numerous DTs. It utilizes the feature bagging technique, where each decision tree is trained on a random subset of features rather than the full feature set. The final prediction is generated by averaging the outputs of all individual trees in the forest. While RFs provide better resistance against overfitting than a single DT, they require higher computational power and memory to store and process the multiple trees. 

\subsubsection*{Gradient Boosted Trees (GBs)} build an ensemble of trees in a sequential manner. At each stage, the model fits a new tree to the residual errors of the current prediction. Each new tree in the sequence is trained to correct the errors made by the preceding ensemble, and the final predictor is obtained by adding the contributions of all trees. This stage-wise optimization allows GB to model nonlinear relationships and makes it useful for structured tabular data.
GBs give high performance but can be sensitive to hyperparameter tuning. 
Our experiments revealed that GBs achieved the best performance on the Aurora and Frontier systems. Therefore, we use them in our active and generative learning frameworks.

\subsubsection*{Ada Boosting Trees (ABs)} like GBs, sequentially construct a series of base learners, such as DTs, focusing on data points that were previously mispredicted. The difference from GBs is that ABs dynamically adjust the weights of both the data points and the individual learners throughout the training process~\cite{freund1997decision}. A vulnerability of ABs is their sensitivity to noisy data and outliers. Because ABs exponentially increase the weight of mispredicted samples, the model can overemphasize noisy data or outliers, leading to overfitting.

\subsubsection*{Gaussian Processes (GPs)} are probabilistic models that define a prior distribution over functions using a kernel. Since GPs are Bayesian, their predictions include uncertainty estimates through posterior means and variance~\cite{rasmussen2006gaussian}. These uncertainty estimates are useful for obtaining prediction confidence, which is essential for active learning methods such as uncertainty sampling.
However, they show poor scalability because the computational complexity grows cubically with the size of the training dataset. 

\subsubsection*{Bayesian Ridge Regression (BR)} enhances the standard ridge regression model by incorporating Bayesian inference. It establishes prior distributions on the model coefficients and then calculates their posterior distributions. Compared with standard ridge regression, BR treats model parameters as random variables with probability distributions, rather than fixed point estimates. This can make the model more stable when the input features are correlated.  BR has the capacity to automatically determine the optimal regularization parameters from the data by maximizing the marginal likelihood~\cite{bishop2006pattern}.

\subsubsection*{Support Vector Regression (SVR)} is the regression counterpart to Support Vector Machines (SVM). Its objective is to find a function that best approximates the data by fitting a tube around it, minimizing the prediction error for points outside this margin. SVR identifies support vectors, which are data points closest to the hyperplane or decision boundary to define the model, rather than treating all points equally. Thus it is less impacted by outliers. As with KRR, the the performance is influenced by the kernel function and hyperparameters. SVR can perform well on nonlinear regression problems, but its computational cost can become high when the training dataset is large.

\subsubsection*{Deep Neural Networks (DNNs)} are structured with multiple layers of interconnected nodes, or neurons. Each neuron performs a computation by applying a non-linear activation function to a weighted sum of its inputs~\cite{goodfellow2016deep}. For the scope of this work, DNNs were not employed, as the traditional algorithms discussed above provided sufficient predictive accuracy while being computationally less expensive.

\subsection{Hyperparameter tuning}
To determine the most effective configuration for our machine learning models, we compared three strategies for hyperparameter tuning: grid search, random search, and Bayesian optimization. We implemented these search algorithms using the GridSearchCV and RandomizedSearchCV from Scikit-learn~\cite{sklearn} and the BayesSearchCV class from the Scikit-optimize library~\cite{skopthead_2022_6451894}. 
Grid Search: A brute-force method that evaluates all combinations in a parameter grid. For $m$ parameters with $n_j$ values each, it tests $\prod_{j=1}^{m} n_j$ configurations. This becomes computationally expensive with many hyperparameters.

Random Search: Samples hyperparameters randomly from defined distributions, significantly reducing computation by limiting evaluations. It is effective when only a few parameters strongly influence model performance~\cite{random10.5555/2188385.2188395}.

Bayesian Search: Models the objective function (e.g., performance metric) using Gaussian Processes, updating the parameter distribution after each evaluation. An acquisition function selects the next promising configuration. This adaptive approach typically finds optimal settings with fewer evaluations~\cite{snoek2012practicalbayesianoptimizationmachine}.

\subsubsection*{Evaluation metrics}
The $R^2$ score, or the coefficient of determination, is a statistical metric that quantifies the proportion of variance in the dependent variable with the change in independent variable. The value of $R^2$ ranges from 0 to 1. Values closer to 1 indicate that the model accounts for a larger portion of the data's variability, signifying a better fit. The formula is as follows:
$$R^2 = 1 - \frac{\sum_{i=1}^{n} (y_i - \hat{y}_i)^2}{\sum_{i=1}^{n} (y_i - \bar{y})^2}$$
Here, $y_i$ represents the series of observed data points, $\hat{y}_i$ are the corresponding values predicted by the model, and $\bar{y}$ is the mean of the observed data.

The Mean Absolute Error (MAE) is a scale-dependent metric that calculates the average of the absolute differences between the predicted and actual values. It provides a measure of the average magnitude of the prediction errors, irrespective of their direction. It is defined by the equation:
\begin{equation}\nonumber
\text{MAE} = \frac{1}{n}\sum_{i=1}^{n}\left| y_i - \hat{y}_i \right|
\end{equation}
In this formula, $n$ denotes the total number of data points, $y_i$ are the actual values, and $\hat{y}_i$ are the model's predictions.

MAPE is a scale-independent metric that expresses error as a percentage of the observed value. When data magnitudes vary widely, focusing on percentage error provides a clearer understanding of performance because it evaluates the error in proportion to each actual value. It is calculated as:
\begin{equation}\nonumber
\text{MAPE} = \frac{1}{n}\sum_{i=1}^{n}\left|\frac{y_i - \hat{y}_i}{y_i}\right| \times 100\%
\end{equation}

For both MAE and MAPE, lower values suggest superior model performance. Our methodology incorporated an integrated k-fold cross-validation process via Scikit-learn and Skopt. During model tuning, the optimization process was configured to either maximize the $R^2$ score or minimize the negative mean absolute percentage error.

\subsection{Baseline approach}
Our framework is designed to assist application users in making decisions about computational resources by addressing two real-world cases.
The first case is when generating new data points is not a limiting factor. In this context, we perform a sufficient number of CCSD calculations on a target supercomputing platform. From these runs, we collect data points for a specific set of runtime parameters (including problem size, node counts, and tile sizes) and their total execution times, and then use this dataset to train a hyperparameter-optimized ML model. Our baseline approach consists of evaluating a range of traditional, non-deep learning ML models. With this approach, we aim to answer two types of user queries:

\subsubsection*{Shortest-time question (STQ)}
This question asks to identify the exact runtime parameter configuration that will result in the fastest execution for a given problem on a particular supercomputer. One example of an STQ is "What is the smallest possible execution time for a CCSD computation with a problem size of (O, V) on the Aurora supercomputer?"

\subsubsection*{Budget question (BQ)} 
The Budget Question (BQ) asks, given a finite allocation of resources—whether measured in time in seconds, node-seconds, or electricity cost—what experimental configurations are actually feasible? Our work, for example, addresses queries such as "What combinations of problem size (O, V), node count, and tile size are possible for a CCSD run on Aurora under a 1000-second time limit?"

\begin{figure}[t]
    \centering
    \includegraphics[width=0.7\linewidth]{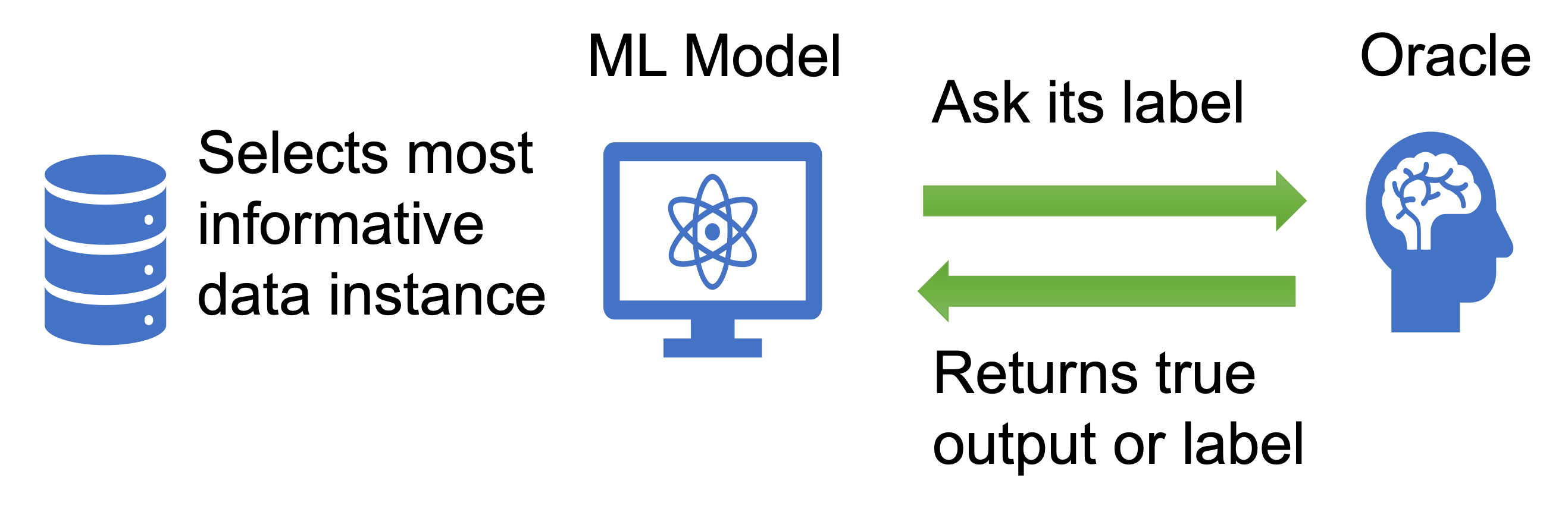}
    \vspace{-\bigskipamount}
    \caption{Active learning.}
    \label{fig:active_learning}
\end{figure}
\subsection{Active learning}
The second real-world scenario we consider is driven by the scarcity of an application's historical data on a supercomputer, which is expensive to run. Since there is insufficient data, we employ Active Learning strategies to demonstrate that a model can be effectively trained even with a small dataset.
Active learning techniques differ from traditional ML models in that they do not train on a static dataset. This is an interactive process where the active learning algorithm selects the most informative unlabeled data points to be labeled by an information source or ``oracle.'' By identifying the most informative data points, active learning reduces the number of labeled data points required to train a model. 
This is especially useful in supervised learning, where runs required to acquire labeled data are either slow or expensive, leading to inadequate data. We use active learning to select the most informative CCSD parameter configurations. Because supercomputer trial runs are costly, our goal is to answer the STQ/BQ with as few training data as possible. Figure \ref{fig:active_learning} visualizes active learning.

There are many query strategies in active learning.
In Uncertainty Sampling (US) (Algorithm~\ref{algorithm:us}), the learner queries the points for which its current model is least certain about the predicted outputs.
In Query-by-Committee (QC) (Algorithm~\ref{algorithm:qc}), we query points that show the most disagreement among a committee of models trained on the available data. 
Other active learning strategies include expected model change, where query points are predicted to cause the largest improvement in model parameters, and expected error reduction, where query points are expected to reduce generalization error the most.
We use Gaussian Processes (GPs) for US and Gradient Boosting (GB) for QC.
When our objective is STQ, as opposed to ML model assessment, we identify the true and predicted optimal parameters from the execution times in both our ground truth test dataset and predicted execution times produced by the model. To evaluate loss, we first identify the configuration that is optimal under the predicted total execution time, then measure performance using the true execution time achieved by that configuration, rather than its predicted time. 
Losses computed from the predicted time are, in most cases, smaller than those based on true time from the predicted optimal configurations and thus underestimate the true loss. To ensure accurate model evaluation, we instead identify the predicted parameter configuration and use its corresponding true execution time to compute the loss.

\begin{algorithm}[t]
\caption{Uncertainty Sampling (US) with Gaussian Processes}
\label{algorithm:us}
\begin{algorithmic}
\STATE $n_{\text{initial}} \leftarrow 50$
\STATE $\text{indices} \leftarrow \text{Randomly select } n_{\text{initial}} \text{ indices from } X_{\text{train}}$
\STATE $X_{\text{labeled}} \leftarrow X_{\text{train}}[\text{indices}]$;
$y_{\text{labeled}} \leftarrow y_{\text{train}}[\text{indices}]$
\STATE $X_{\text{unlabeled}} \leftarrow X_{\text{train}} \setminus 
 X_{\text{labeled}}$;
$y_{\text{unlabeled}} \leftarrow y_{\text{train}} \setminus 
 y_{\text{labeled}}$
\STATE $n_{\text{queries}} \leftarrow 20$;
\STATE $\text{query\_size} \leftarrow 50$
\STATE $is\_STQ \leftarrow \textbf{True/False} $
\IF{$is\_STQ$}
\STATE $\text{optimal\_test} = \text{get\_optimal\_values}(X_{\text{test}}, y_{\text{test}})$
\ENDIF
\STATE $\text{model} \leftarrow \text{GaussianProcessRegression(...)}$
\FOR{$i = 1 \to n_{\text{queries}}$}
    
    \STATE $ \_, \text{std}\leftarrow \text{model.predict}(X_{\text{unlabeled}},\text{return\_std=True})$
    \STATE $\text{q\_indices} \leftarrow \text{argsort}(-\text{std})[:\text{query\_size}]$
    \STATE $X_{\text{labeled}} \leftarrow X_{\text{labeled}} \cup X_{\text{unlabeled}}[\text{q\_indices}]$
    \STATE $y_{\text{labeled}} \leftarrow y_{\text{labeled}} \cup y_{\text{unlabeled}}[\text{q\_indices}]$
    \STATE $X_{\text{unlabeled}} \leftarrow X_{\text{unlabeled}} \setminus X_{\text{unlabeled}}[\text{q\_indices}]$
    \STATE $y_{\text{unlabeled}} \leftarrow y_{\text{unlabeled}} \setminus y_{\text{unlabeled}}[\text{q\_indices}]$
    \IF{$is\_STQ$}
    \STATE $y_{\text{pred}} \leftarrow \text{model.predict}(X_{\text{test}})$
    \STATE $\text{optimal\_pred} \leftarrow \text{get\_optimal\_values}(X_{\text{test}}, y_{\text{pred}})$
    \STATE $\text{r2, mae, mape} \leftarrow ...$ 
    \STATE $... \text{compute\_{losses} ($X_{\text{test}}$,  $y_{\text{test}}$  optimal\_test, optimal\_pred )}$
    
    \ENDIF
\ENDFOR
\end{algorithmic}
\end{algorithm}
\begin{algorithm}[t]
\caption{Query-by-Committee (QC) with Gradient Boosting}
\label{algorithm:qc}
\begin{algorithmic}
\STATE $n_{\text{initial}} \leftarrow 50$
\STATE $\text{indices} \leftarrow \text{Randomly select } n_{\text{initial}} \text{ indices from } X_{\text{train}}$
\STATE $X_{\text{labeled}} \leftarrow X_{\text{train}}[\text{indices}]$;
 $y_{\text{labeled}} \leftarrow y_{\text{train}}[\text{indices}]$
\STATE $X_{\text{unlabeled}} \leftarrow X_{\text{train}} \setminus X_{\text{labeled}}$;
 $y_{\text{unlabeled}} \leftarrow y_{\text{train}} \setminus y_{\text{labeled}}$
\STATE $n_{\text{committees}} \leftarrow 5$; 
 $n_{\text{queries}} \leftarrow 10$
\STATE $\text{query\_size} \leftarrow 50$
\STATE $is\_STQ \leftarrow \textbf{True/False} $
\IF{$is\_STQ$}
\STATE $\text{optimal\_test} = \text{get\_optimal\_values}(X_{\text{test}}, y_{\text{test}})$
\ENDIF
\FOR{$i = 1 \to n_{\text{queries}}$}
    \STATE $\text{committee\_predictions} \leftarrow ....$
    \STATE $... \text{zeros}(X_{\text{unlabeled}}.\text{shape}[0], n_{\text{committees}})$
    \FOR{$j = 1 \to n_{\text{committees}}$}
        \STATE $\text{model} \leftarrow 
        \text{GradientBoostingRegressor}(...)$
        \STATE $\text{model.fit}(X_{\text{labeled}}, y_{\text{labeled}})$
        \STATE $\text{committee\_predictions}[:, j] \leftarrow \text{model.predict}(X_{\text{unlabeled}})$
    \ENDFOR
    \STATE $\text{prediction\_variance} \leftarrow \text{{var}}(\text{committee\_predictions})$
    \STATE $\text{q\_indices} \leftarrow \text{{argsort}}(-\text{prediction\_variance})[:\text{query\_size}]$
    \STATE $X_{\text{query}}, y_{\text{query}} \leftarrow X_{\text{unlabeled}}[\text{q\_indices}], y_{\text{unlabeled}}[\text{q\_indices}]$
    \STATE $X_{\text{labeled}} \leftarrow \text{vstack}([X_{\text{labeled}}, X_{\text{query}}])$
    \STATE $y_{\text{labeled}} \leftarrow \text{append}(y_{\text{labeled}}, y_{\text{query}})$
    \STATE $X_{\text{unlabeled}} \leftarrow X_{\text{unlabeled}} \setminus X_{\text{query}}$
    \STATE $y_{\text{unlabeled}} \leftarrow y_{\text{unlabeled}} \setminus y_{\text{query}}$
    
    \IF{$is\_STQ$}
    \STATE $y_{\text{pred}} \leftarrow \text{model.predict}(X_{\text{test}})$
    \STATE $\text{optimal\_pred} \leftarrow \text{get\_optimal\_values}(X_{\text{test}}, y_{\text{pred}})$
    \STATE $\text{r2, mae, mape} \leftarrow ...$
    \STATE $...\text{compute\_losses} (X_{\text{test}}, y_{\text{test}}, \text{optimal\_test, optimal\_pred})$
    
    \ENDIF
\ENDFOR
\end{algorithmic}
\end{algorithm}

\subsection{Generative learning}
\label{sub:generative-learning}
Similar to active learning,
when training data is scarce, synthetic data generation with generative learning can enhance prediction performance by augmenting the measured dataset with synthetic samples that preserve key statistical data properties of the original data. We use two well-known generative learning methods for single tables: Gaussian Copula~\cite{SDV} and Conditional Tabular Generative Adversarial Networks (CTGAN)~\cite{CTGAN2019}. We use both synthesizers to evaluate whether a faster statistical generator is enough to improve runtime prediction, or whether a neural generative model provides additional benefit.

The Gaussian Copula Synthesizer models dependencies between variables by transforming data into a Gaussian space and capturing their correlations via a copula function. More specifically, the synthesizer first learns the marginal distribution of each column so that variables with different scales and distributional shapes can be modeled separately. It then normalizes the values into a Gaussian representation and learns the covariance between pairs of normalized columns.The synthesizer then samples new values in the Gaussian representation and converts them back to the original column formats, producing synthetic rows with the same variables as the measured data. It is fast and well-suited for tabular datasets with continuous variables and monotonic relationships. 

Conditional Tabular GAN (CTGAN) is a deep generative model designed to handle mixed-type tabular data. It uses conditional sampling and mode-specific normalization to generate realistic samples that preserve complex interactions, including categorical patterns. CTGAN addresses several challenges common in tabular data, including mixed discrete and continuous columns, non-Gaussian and multimodal continuous variables, and imbalanced discrete values. Mode-specific normalization is used for continuous variables, while conditional generation and training-by-sampling improve learning from imbalanced discrete columns~\cite{CTGAN2019}. Since our runtime data include discrete configuration parameters such as node count and tile size, with some values appearing much less often than others, CTGAN is a useful comparison to Gaussian Copula for synthetic data generation.

Both models have gained traction in domains requiring privacy-preserving data sharing, performance modeling, and training augmentation. In our work, we explore their application to augment sparse training data collected from supercomputer runs to enhance the accuracy of our ML models for runtime and resource prediction.

\section{Experimental evaluation}
\label{experimental_evaluation}

This section evaluates the proposed approaches using CCSD runtime data from Aurora, Frontier, and Perlmutter. After describing the datasets and experimental setup, we compare the candidate models and select the model which is then used in the subsequent experiments. The evaluation then moves from the supervised baseline to cases where only a small number of measured runs are available, where active learning and generative learning are used to reduce the need for additional CCSD runs. Throughout the section, we focus on two practical decision problems: identifying the configuration with the shortest runtime and identifying configurations that satisfy a resource budget.

\subsection{Methodology and experimental setup}
We generated our dataset using the high-performance tensor algebra framework Tensor Algebra for Many-body Methods (TAMM)~\cite{tamm_jcp_2023}, which provides scalable and portable implementations of many-body methods such as CCSD~\cite{ccsd1,ccsd2,ccsd3}. TAMM is a distributed tensor algebra library designed for modern supercomputing architectures. The CCSD calculations were performed using ExaChem~\cite{exachem_2023}, an open-source, scalable electronic structure package built on top of the TAMM framework.

The computational cost of a CCSD calculation is determined by the number of molecular orbitals.
In CCSD, the computational problem size is determined by the number of molecular orbitals. These orbitals are categorized into two sets: occupied (O) orbitals, containing electrons in the reference wavefunction, and virtual (V) orbitals, which represent orbitals that are unoccupied and available to potentially hold electrons.
In quantum chemistry calculations, the total number of molecular orbitals, denoted as $N$, is the sum of the occupied ($O$) and virtual ($V$) orbitals, where $N = O + V$. $N$ is equal to the number of functions in the base set that represent the electronic wavefunction of a molecule. Since CCSD scales approximately as $\mathcal{O}(O^2V^4)$, increasing the number of occupied (O) and virtual (V) orbitals causes rapid growth in the time and space complexity of a calculation. Therefore, the parameters (O, V) define the size of the problem and serve as primary input features for modeling or performance prediction.
For the purpose of performance modeling, we execute only a single iteration of the CCSD algorithm for each configuration. This methodology is justified because the overall computational cost of each iteration is dictated by sextic-scaling tensor contractions, i.e., $\mathcal{O}(O^2V^4)$. Since this rate-limiting step is consistent across every iteration, the runtime, memory use, and parallel efficiency observed in one iteration are a reliable indicator for subsequent iterations.
For our study, each computation run was configured with a distinct problem size ($O$, $V$) and specific execution parameters, including the node count and tensor block dimensions. TAMM handles scheduling and execution of the required tensor operations. It uses its task-based runtime, utilizing MPI for distributed execution across nodes and leveraging vendor-optimized libraries to accelerate performance on GPUs.

All of our CCSD experiments were conducted on three premier DOE supercomputing platforms: Aurora (ALCF), Frontier (OLCF) and Perlmutter (NERSC). On each platform, TAMM was configured with optimized settings to ensure the generation of scalable and consistent performance data. From each run, we collected performance data that were then processed into feature sets to train our ML models. The datasets include runtime parameters such as the number of nodes, total wall time for a single iteration, the time required for individual tensor contractions, and the tensor sizes. 
The runtime data collected were used for ML training, active learning evaluation, and synthetic data generation in Python 3.11.1. The Python environment used pandas 2.3.3, NumPy 2.4.2, SciPy 1.17.0, scikit-learn 1.8.0, scikit-optimize 0.10.2, matplotlib 3.10.8, and SDV 1.33.1. Synthetic data generation used SDV’s GaussianCopulaSynthesizer and CTGANSynthesizer; the environment included CTGAN 0.12.0.
 
Table \ref{tab:datasets} presents the partitioning of our data into training and test sets. We determined the optimal parameter configuration for each test set, i.e., the configuration of nodes and tile sizes that resulted in the least total execution time for a given problem size ($O$,$V$).

\begin{table}[tp]
    \centering
    \caption{Datasets and the corresponding size breakdowns.}
    \label{tab:datasets}
    \begin{tabular}{|l|c|c|c|} \hline
      System   & Total & Train & Test  \\ \hline
      Aurora     & 2329 &  1746 & 583 \\ \hline
      Frontier   & 2454  & 1840 & 614 \\ \hline
      Perlmutter   &  2203 & 1652 & 551 \\ \hline
    \end{tabular}
\end{table}
\begin{table}[tp]
\centering
    \caption{Training and prediction times for Gradient Boosting.}
    \label{tab:trainingtimes}
    \begin{tabular}{|c|c|c|} \hline
      System   & Training & Prediction \\ \hline
      Aurora   & 4.86 s ± 202 ms & 81.1 ms ± 4.68 ms \\ \hline
      Frontier   & 5.06 s ± 205 ms & 95.4 ms ± 18 ms  \\ \hline
      Perlmutter & 4.82 s ± 162 ms & 71.3 ms ± 591 $\mu$s \\ \hline
    \end{tabular}
\end{table}
\subsection{Model results and initial analysis}
\begin{figure*}[tp]
    \centering
    \begin{minipage}{0.45\textwidth} 
        \includegraphics[width=\textwidth]{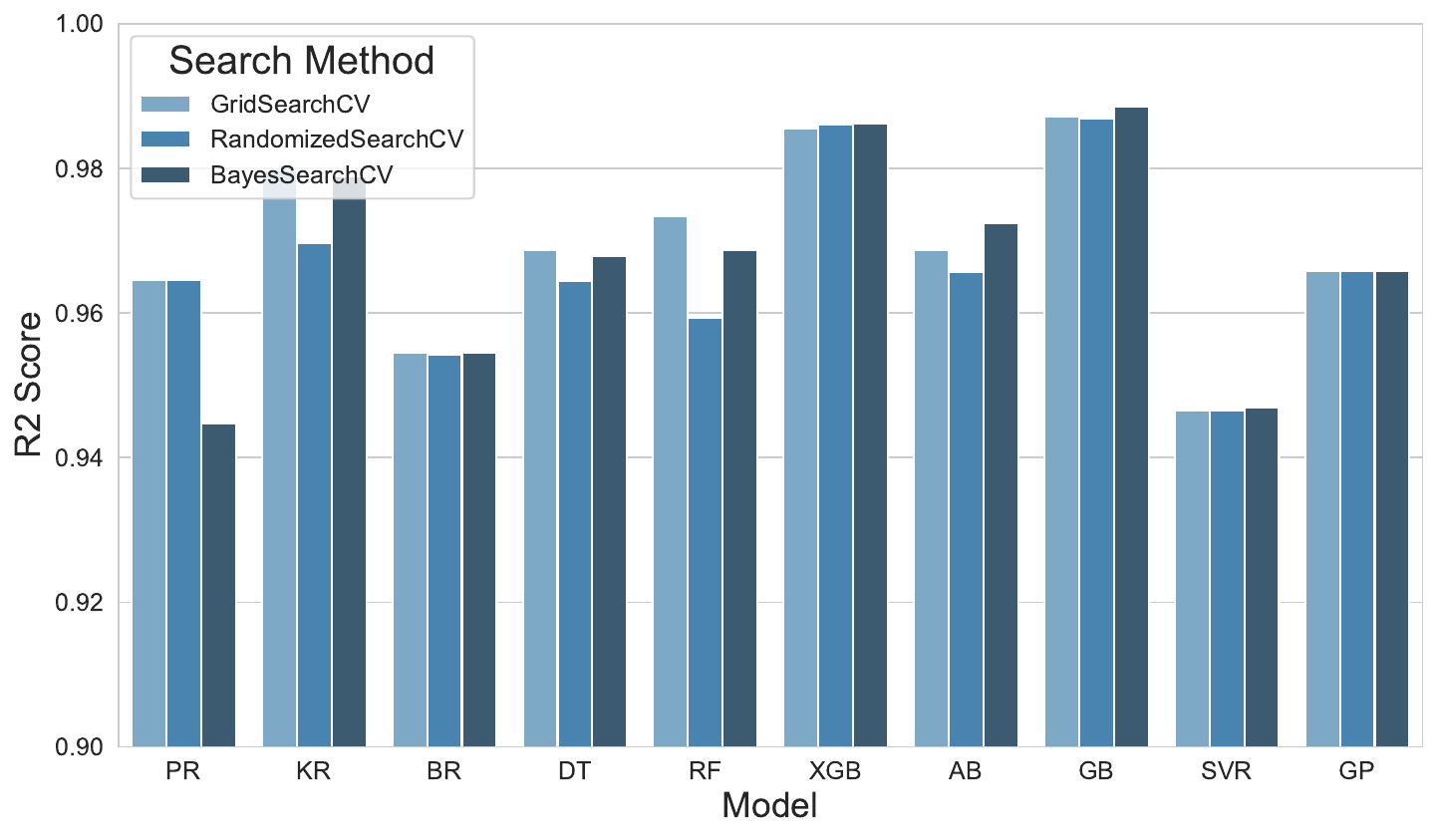}
    \end{minipage}
    \begin{minipage}{0.45\textwidth}
        \includegraphics[width=\textwidth]{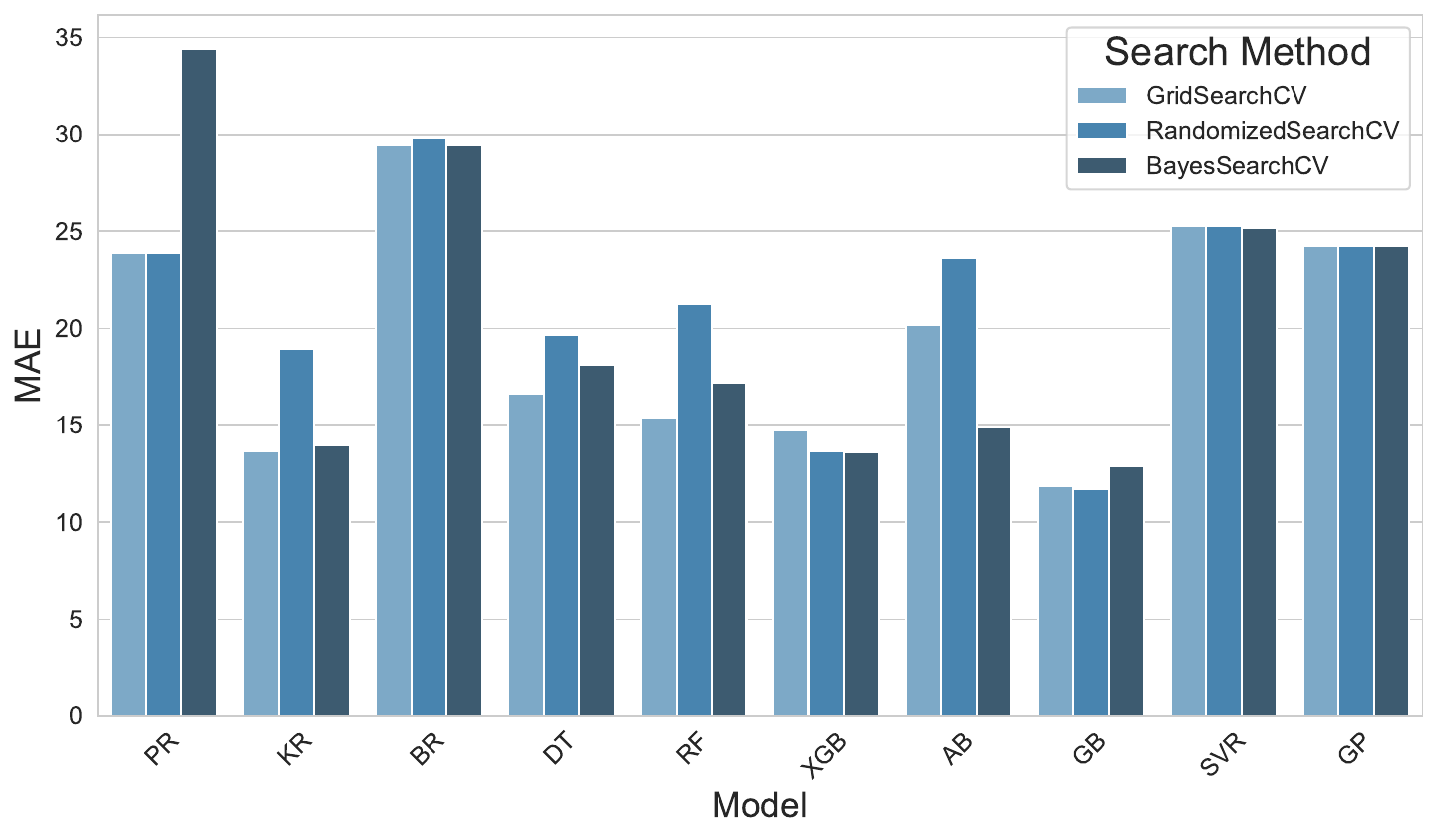}
    \end{minipage}
    \begin{minipage}{0.45\textwidth}
        \includegraphics[width=\textwidth]{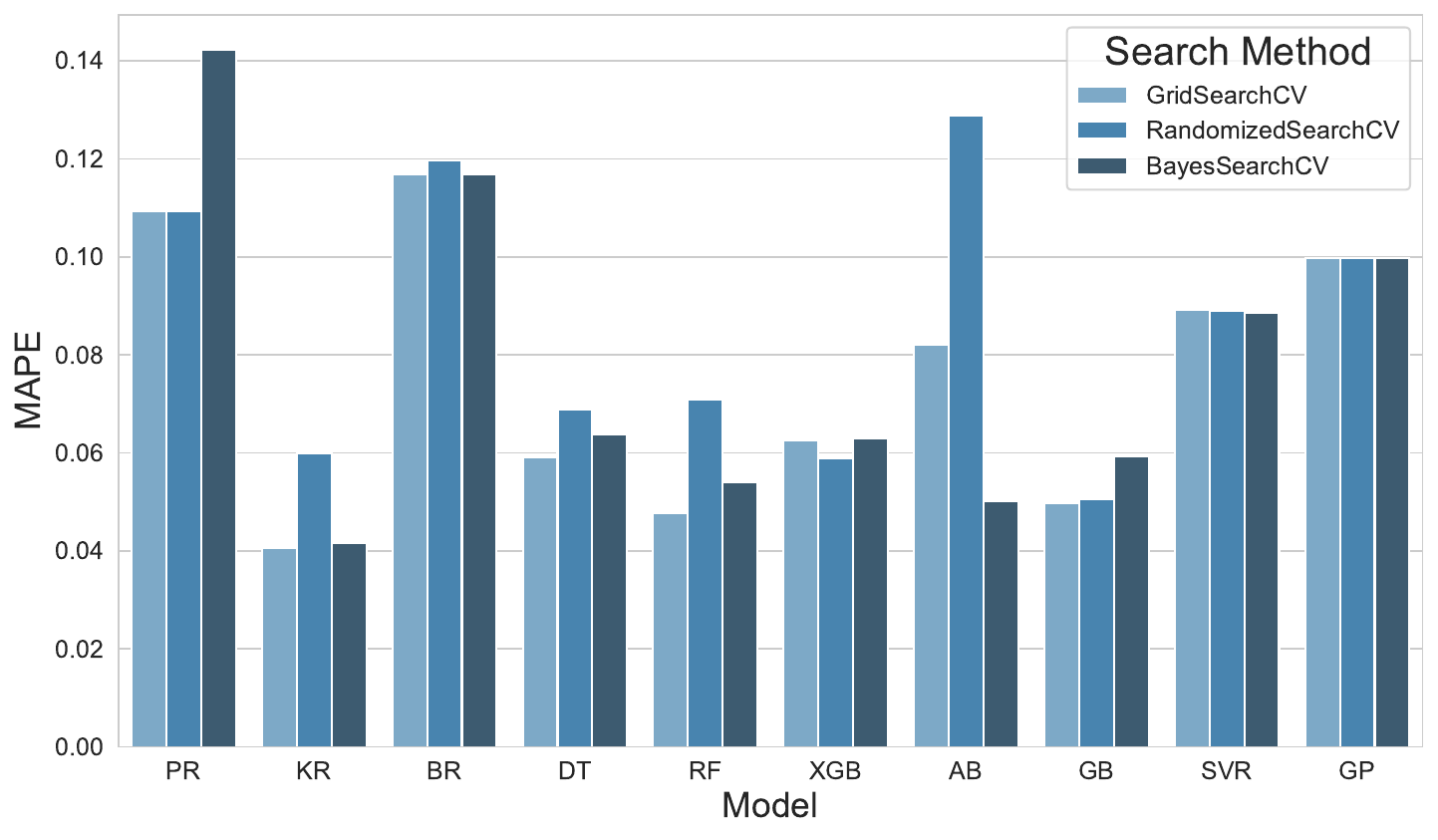}
    \end{minipage}
    \begin{minipage}{0.45\textwidth}
        \includegraphics[width=\textwidth]{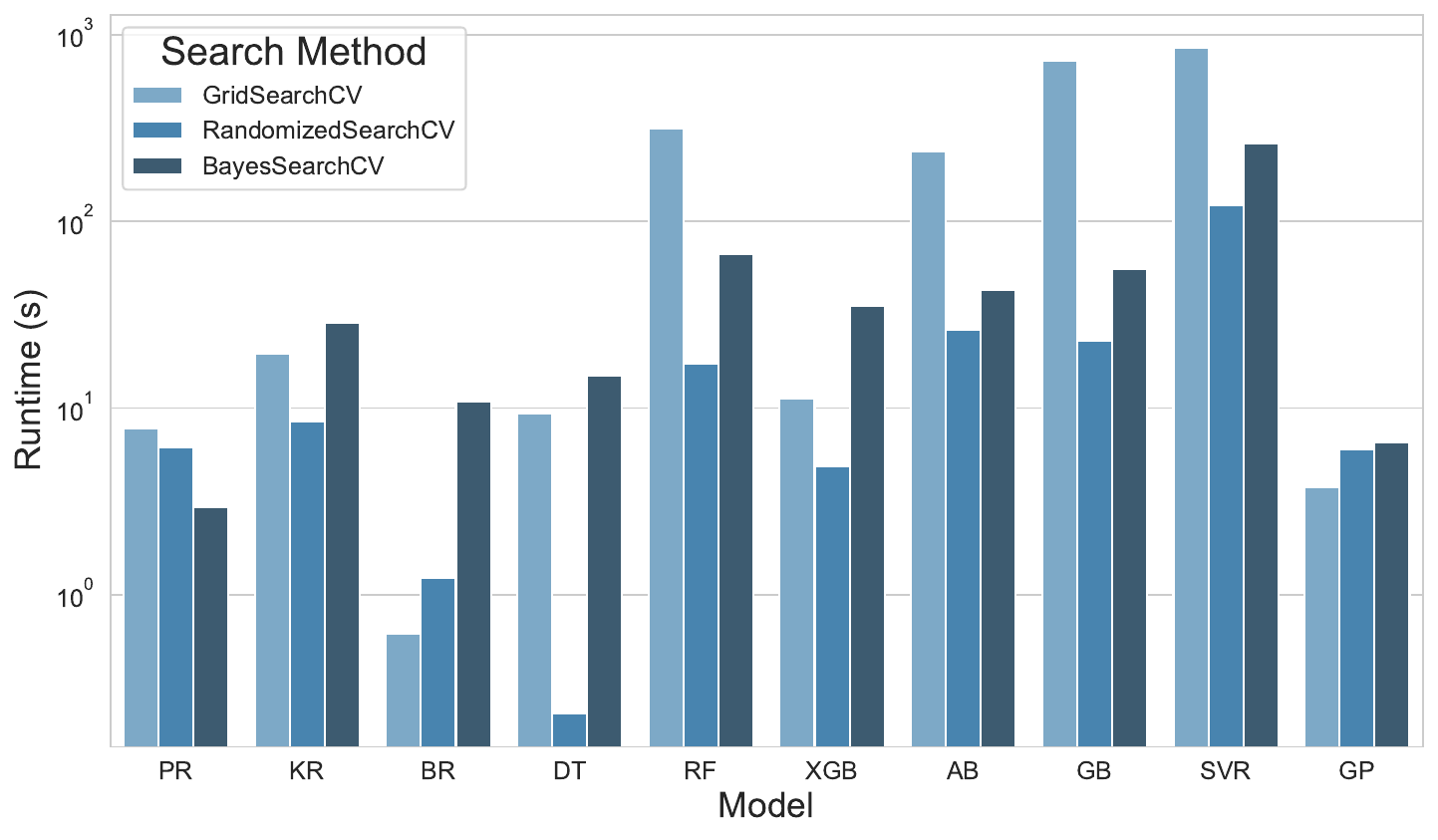}
    \end{minipage}
    \caption{The performances of the selected models with hyperparameter optimization for Aurora}
    \label{fig:aurorahyperparameter}
\end{figure*}

\begin{figure*}[tp]
    \centering
    \begin{minipage}{0.45\textwidth}
        \includegraphics[width=\textwidth]{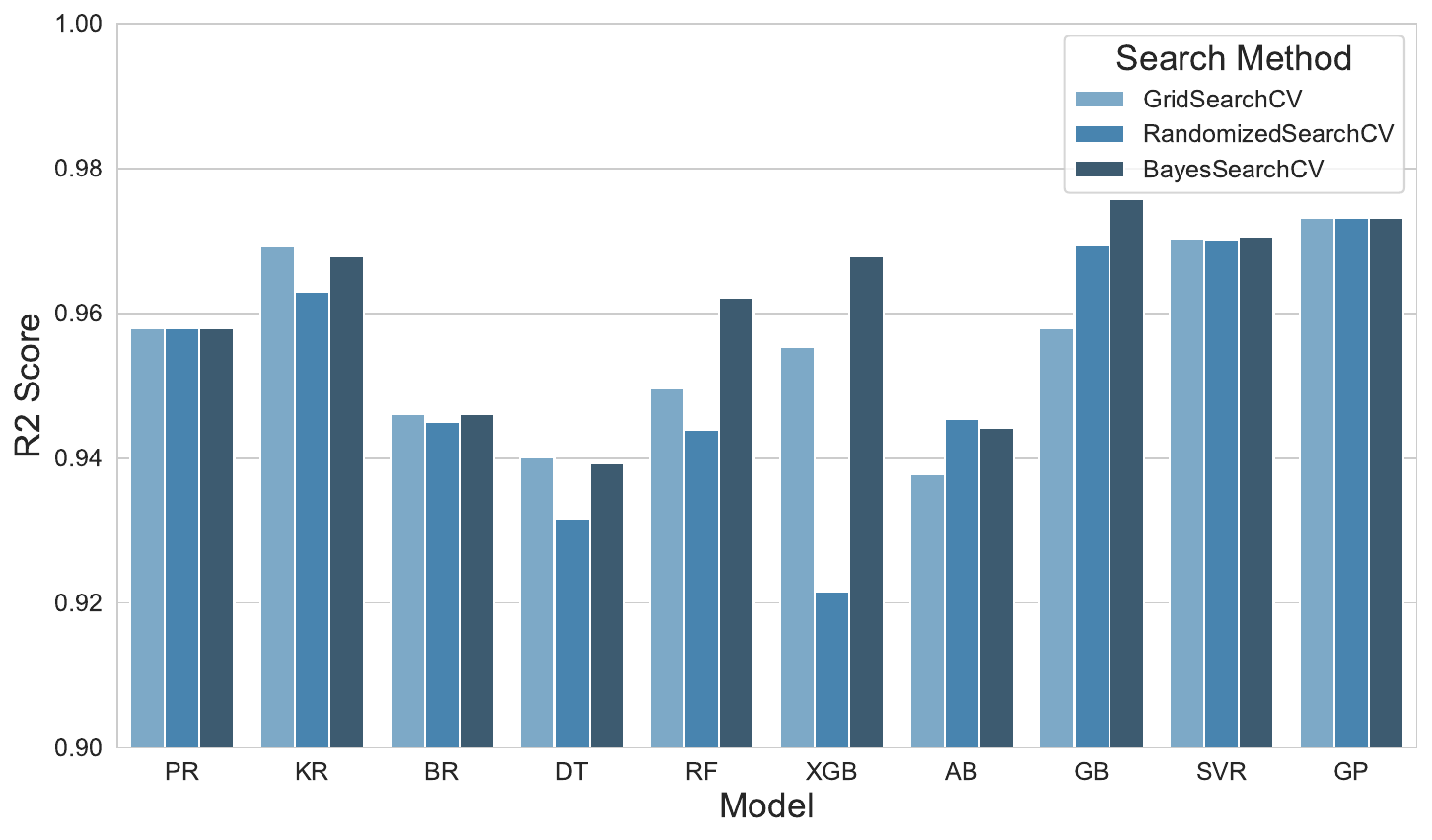}
    \end{minipage}
    \begin{minipage}{0.45\textwidth}
        \includegraphics[width=\textwidth]{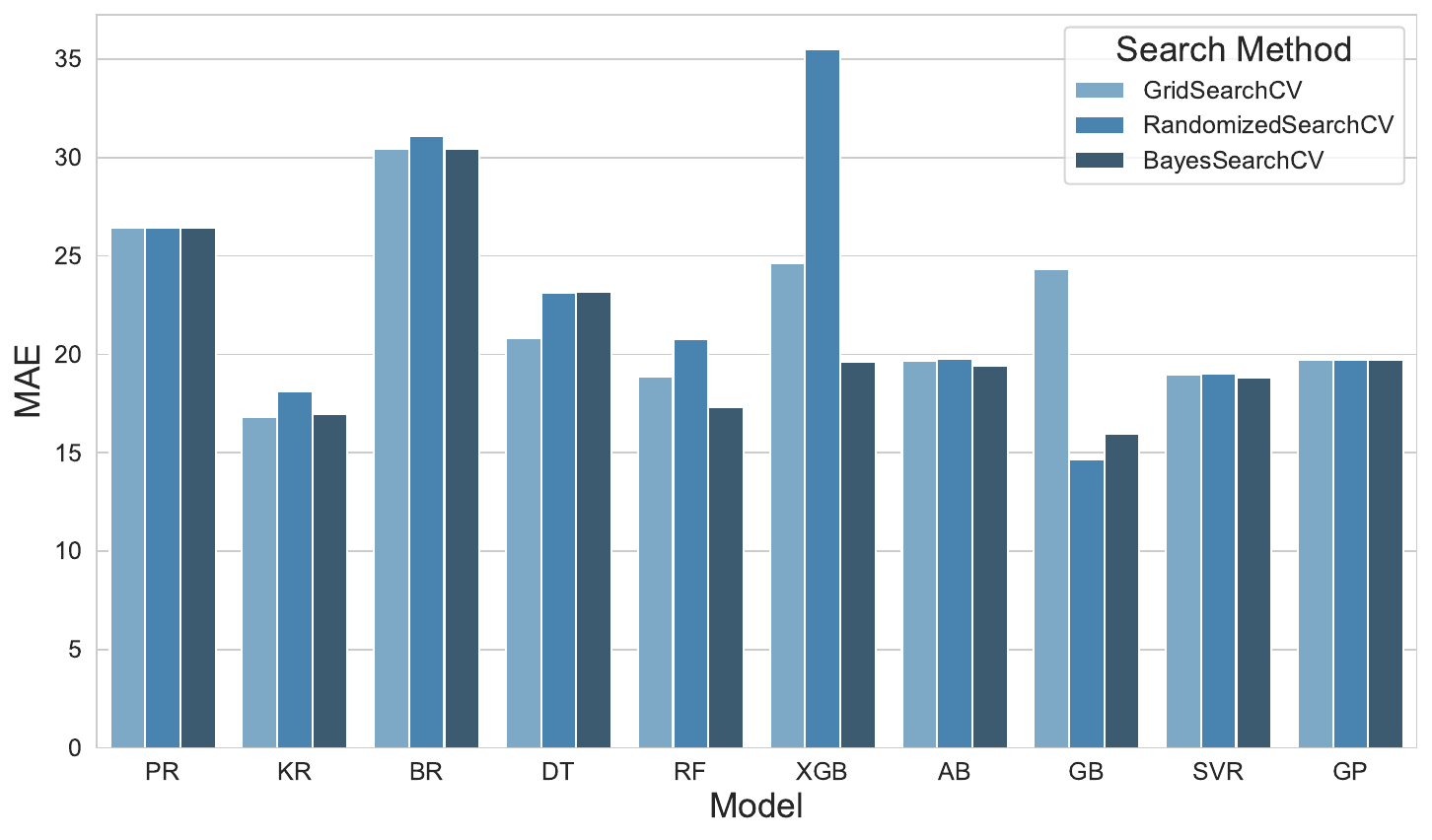}
    \end{minipage}
    \begin{minipage}{0.45\textwidth}
        \includegraphics[width=\textwidth]{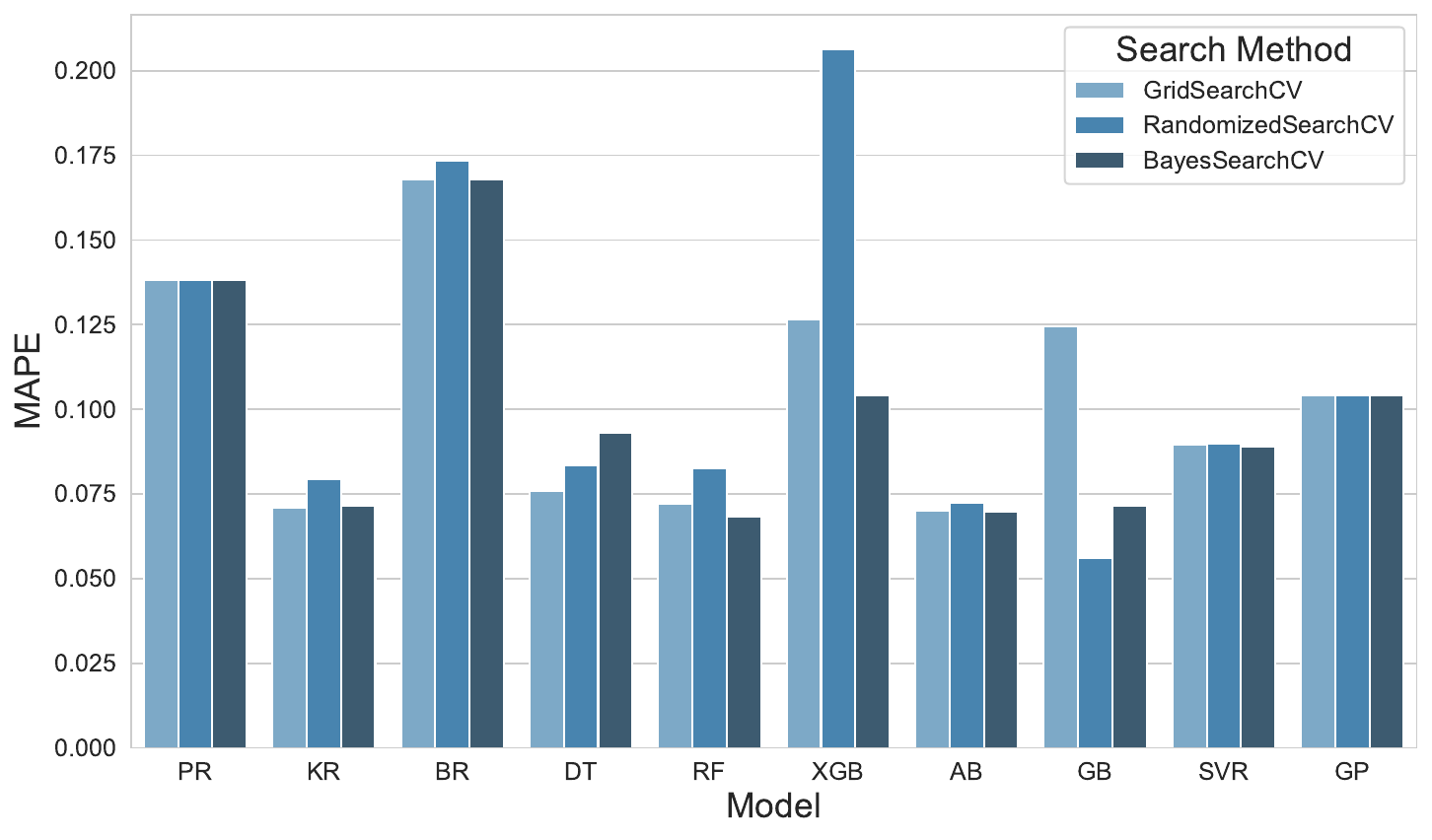}
    \end{minipage}
    \begin{minipage}{0.45\textwidth}
        \includegraphics[width=\textwidth]{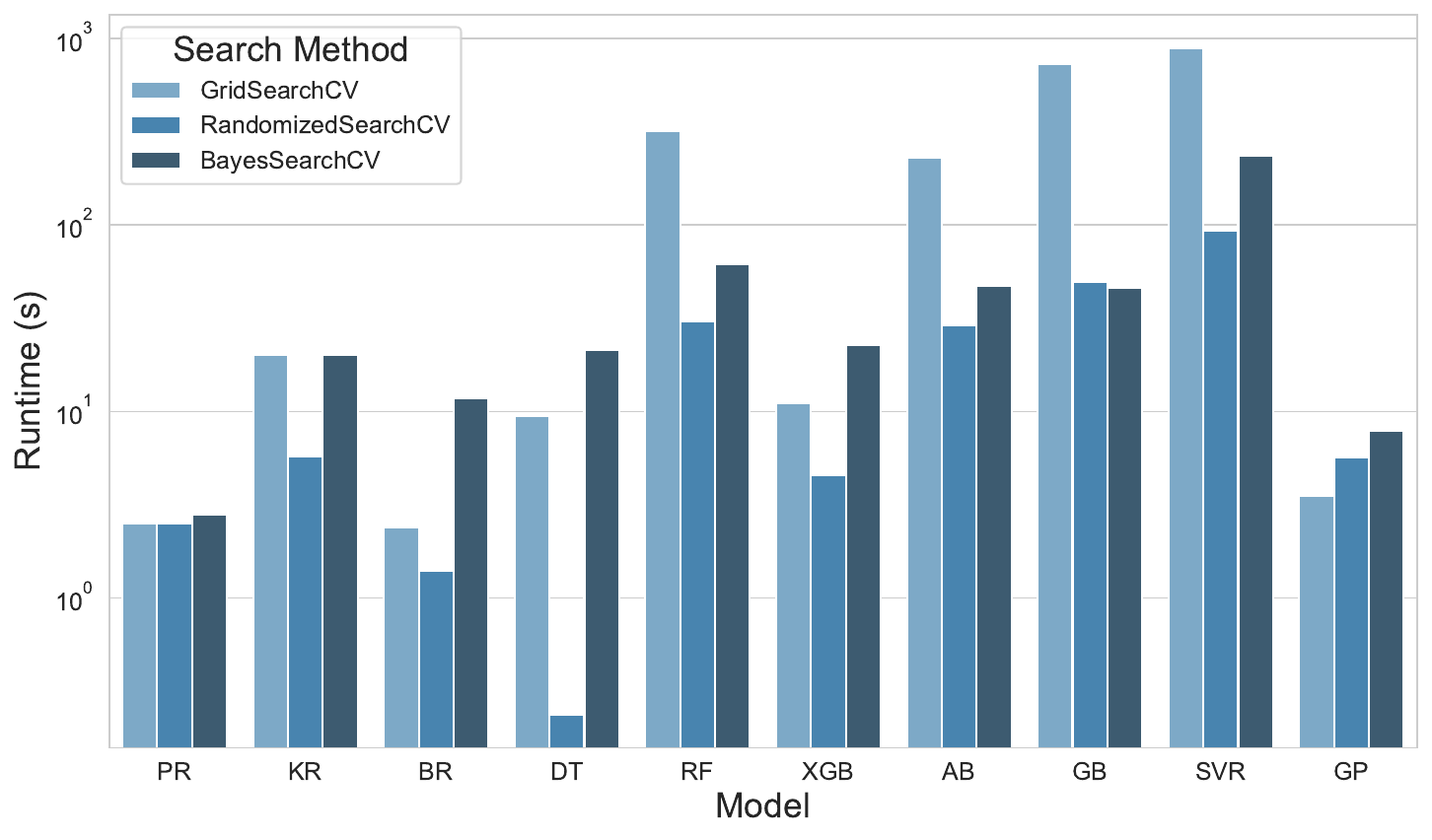}
    \end{minipage}
    \caption{The performances of the selected models with hyperparameter optimization for Frontier}
    \label{fig:frontierhyperparameter}
\end{figure*}

\begin{figure*}[tp]
    \centering
    \begin{minipage}{0.45\textwidth}
        \includegraphics[width=\textwidth]{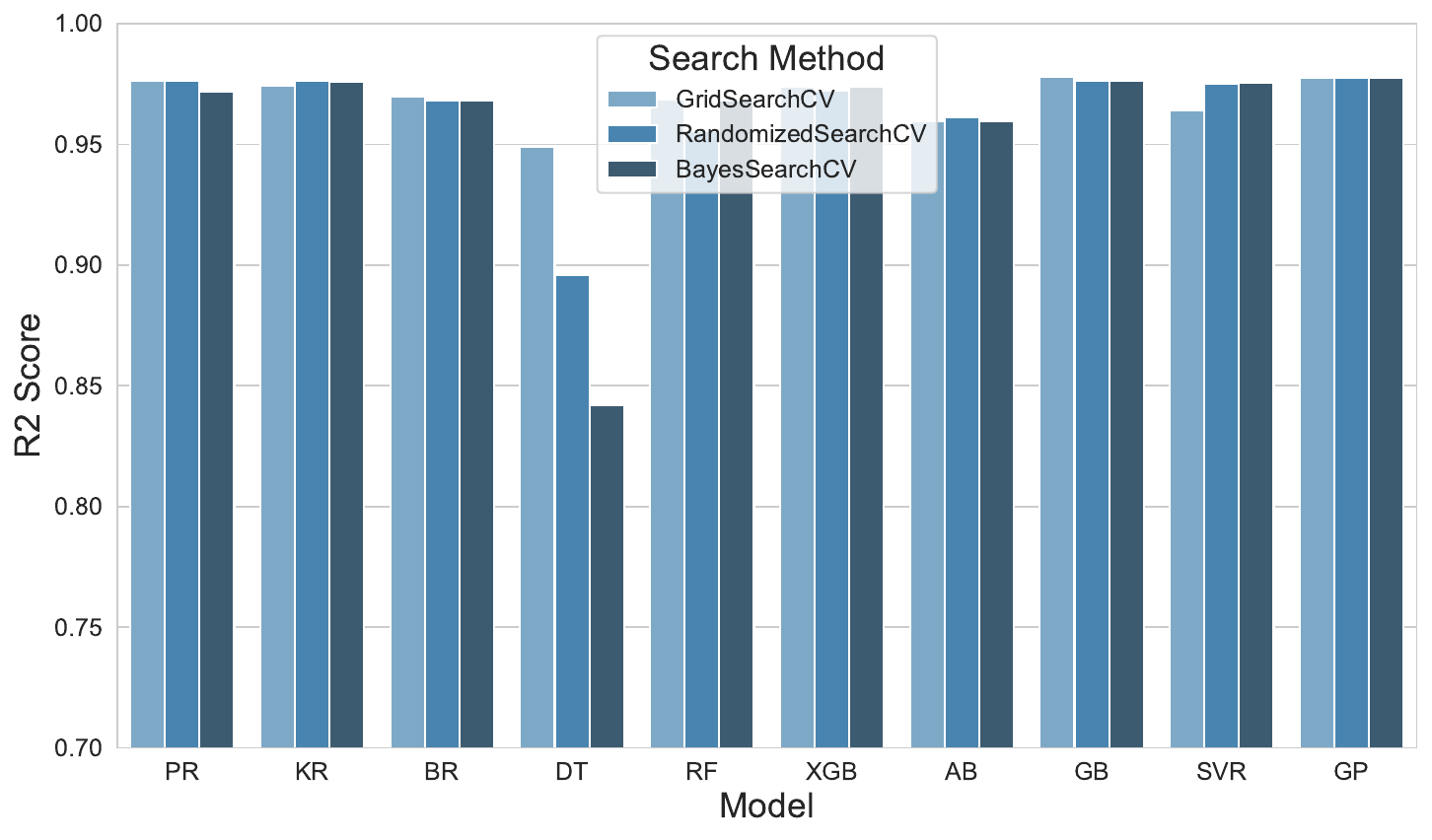}
    \end{minipage}
    \begin{minipage}{0.45\textwidth}
        \includegraphics[width=\textwidth]{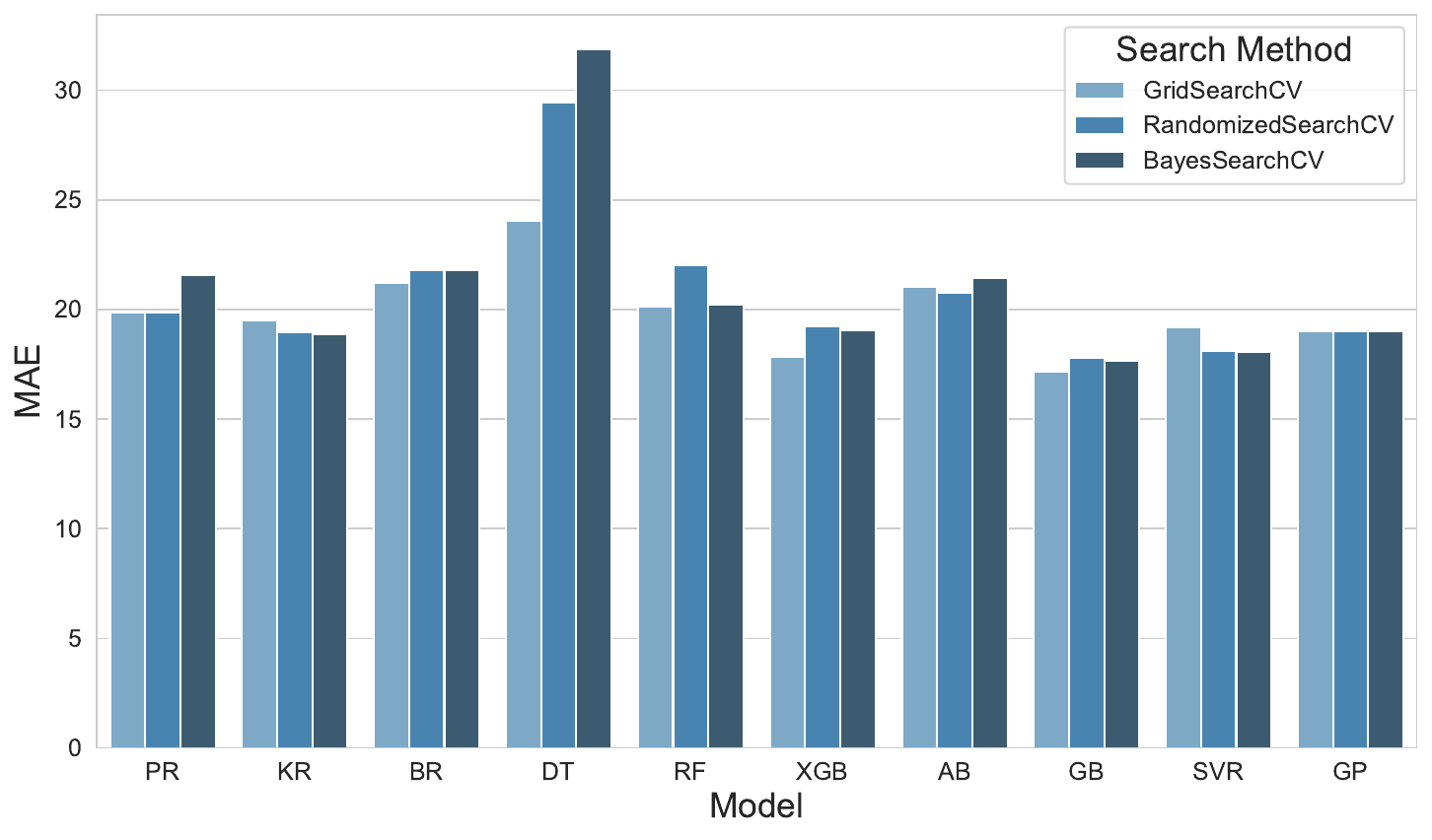}
    \end{minipage}
    \begin{minipage}{0.45\textwidth}
        \includegraphics[width=\textwidth]{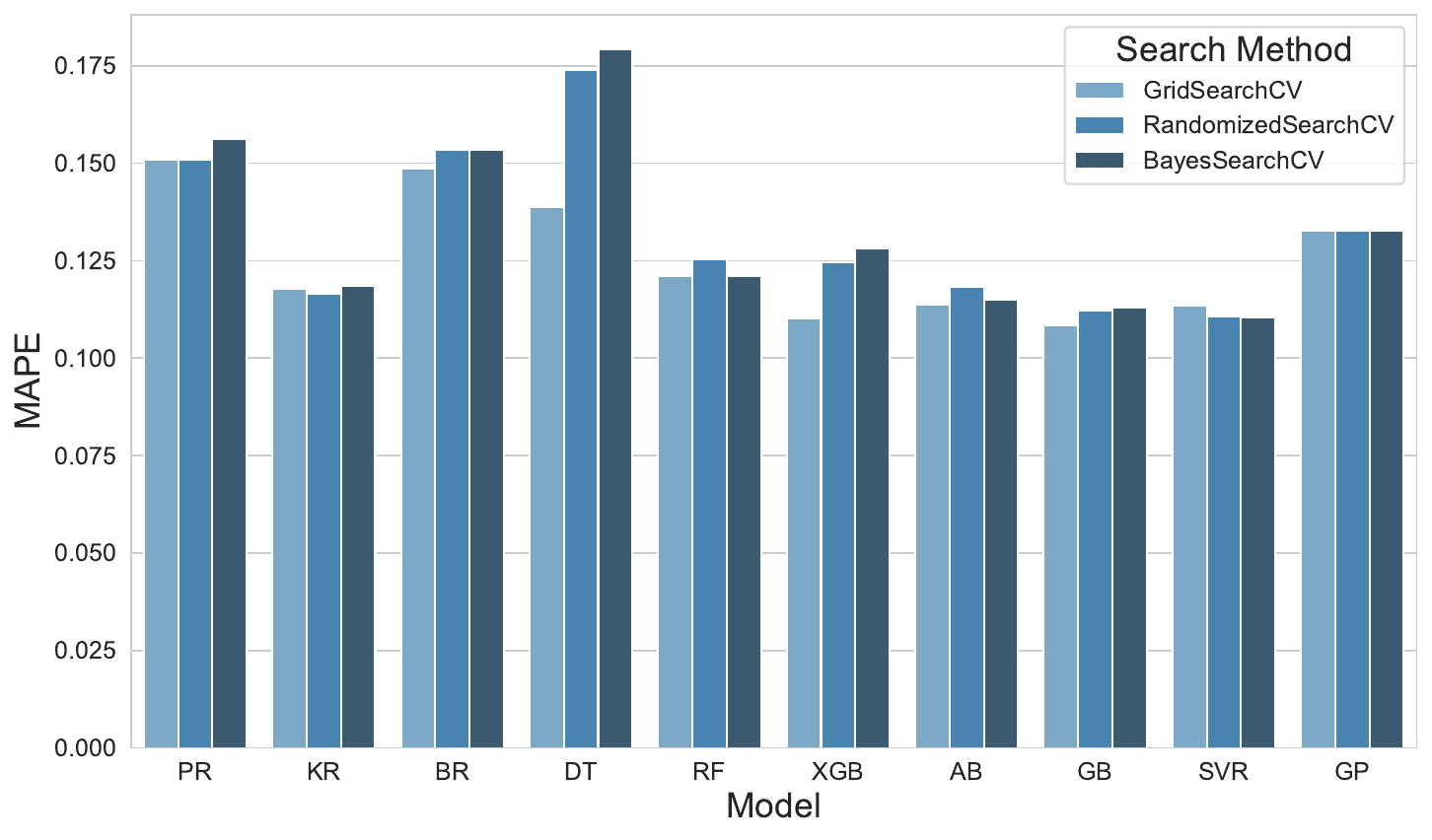}
    \end{minipage}
    \begin{minipage}{0.45\textwidth}
        \includegraphics[width=\textwidth]{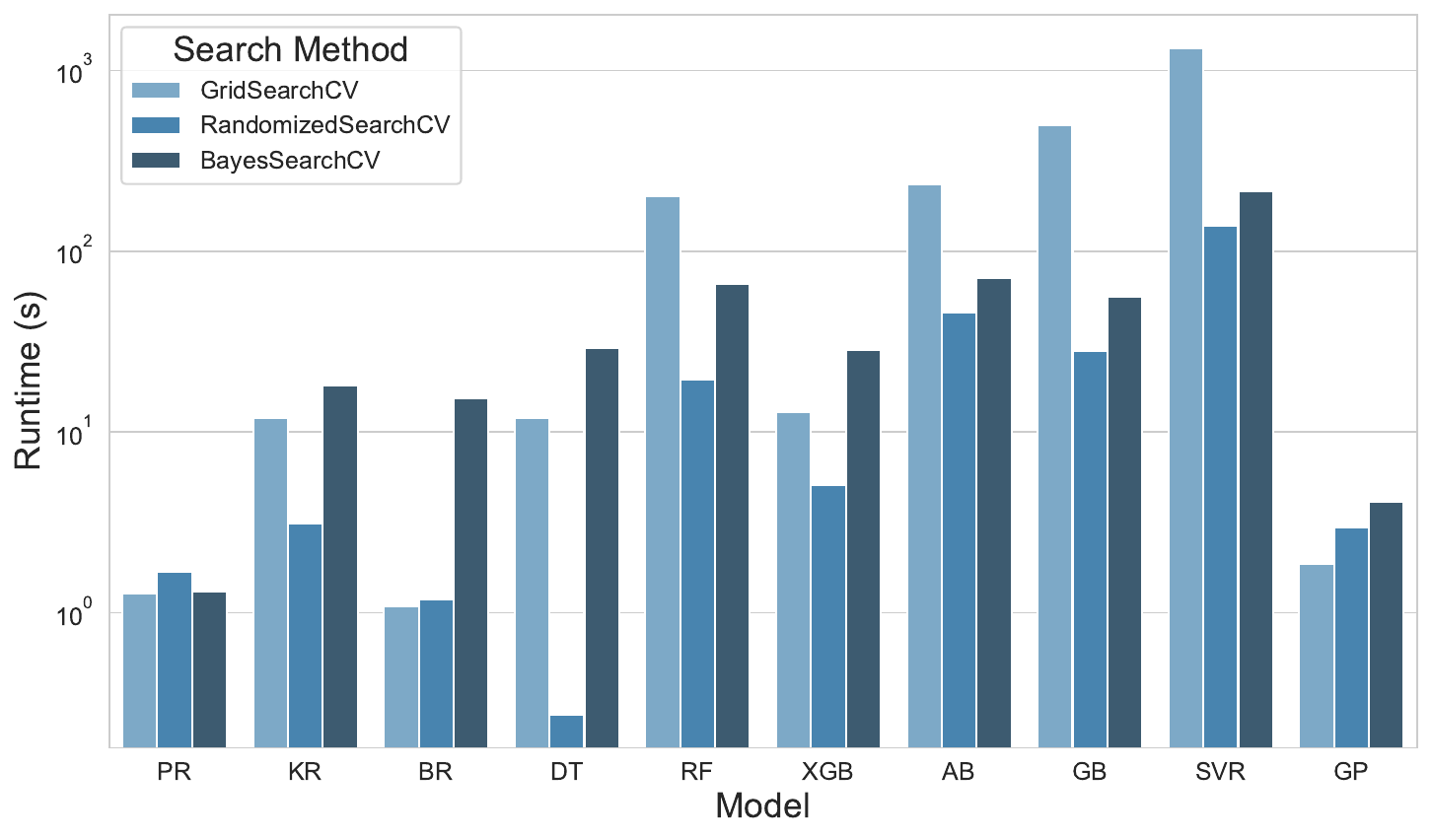}
    \end{minipage}
    \caption{The performances of the selected models with hyperparameter optimization for Perlmutter}
    \label{fig:perlmutterhyperparameter}
\end{figure*}

To achieve the best performance of our ML models, we performed three hyperparameter optimization strategies: grid search, random search, and Bayesian search. Figures \ref{fig:aurorahyperparameter}, \ref{fig:frontierhyperparameter}, and \ref{fig:perlmutterhyperparameter} show the performance of our machine learning models after applying the three hyperparameter tuning strategies for Aurora, Frontier, and Perlmutter, respectively. We compare the resulting evaluation metrics, such as the R² score, MAE, and MAPE, along with the optimization runtimes to determine the best-performing models. 

For all three platforms, the Gradient Boosting (GB) model generally outperformed the others. Thus, we selected GB for our active and generative learning algorithms, except for the Uncertainty Sampling (US) strategy, where we use the Gaussian Process (GP) model. 
Table~\ref{tab:trainingtimes} shows the training and prediction times of a GB model on the Aurora, Frontier, and Perlmutter systems. Training times are close to 5 seconds on all systems, and prediction times vary between about 71 and 96 milliseconds, which are trivial compared to CCSD runtimes. Our hyperparameter optimizations show us that GB tuned with 750 tree-based estimators, a maximum depth of 10, and default hyperparameter values in others performs the best. Hence, we use GB in all our subsequent experiments.
\subsection{Baseline supervised learning results}

In the baseline supervised-learning experiments, the model is trained using the available runtime data and then evaluated on its ability to predict execution times for unseen configurations. These predictions are used to answer the two target questions considered in this work: the Shortest-Time Question (STQ) and the Budget Question (BQ). For STQ, we evaluate whether the model can identify the configuration with the smallest execution time for each problem size (O,V). For BQ, we evaluate whether the model can identify configurations that minimize node-hour cost.

\subsubsection*{Shortest-time question:}
The parameter configurations that produce the shortest execution time for Aurora are highlighted in Table \ref{tab:aurora_shortesttime}. 
In addition, we present the ML model’s predicted values for each $(O, V)$ as explained at the end of Section 3.4.
The model is correct for all cases except those with the predicted execution times shown in parentheses. It incorrectly predicts three parameter configurations and achieves an $R^2$, MAE, and MAPE score of 0.999, 2.36, and 0.023. Even when the predicted optimal configuration is incorrect, the resulting runtime differences from true optimal execution times are negligible.
Similarly, the parameter configurations resulting in the shortest runtimes for Frontier are presented in Table \ref{tab:frontier_shortesttime}, along with our model’s predictions for each problem size $(O, V)$.
Our model makes five incorrect predictions for parameter configurations. Nevertheless, it achieves strong performance with an $R^2$, MAE and MAPE score of 0.979, 3.65, and 0.050.
Finally, Table \ref{tab:perlmutter_shortesttime} shows the parameter configurations leading to the shortest time. It makes fifteen inaccurate predictions. Even with fifteen suboptimal configuration predictions, the $R^2$, MAE and MAPE scores are 0.971, 12.948, and 0.118. 

\subsubsection*{Budget question}
The optimal parameter configurations that give the lowest node-hours for each problem size $(O, V)$ for Aurora are shown in Table \ref{tab:aurora_shortestnodehours}. The values shown inside the parentheses are the incorrect configurations and the corresponding execution times. The model makes correct predictions of the best configurations in all but five cases.  However, the model achieves $R^2$, MAE, and MAPE scores of 0.979, 0.41, and 0.12.
The optimal configurations leading to the smallest node-hours for Frontier are shown in Table \ref{tab:frontier_shortestnodehours}. Here, the model makes suboptimal configuration predictions in nine cases. Despite that, the $R^2$, MAE and MAPE scores are 0.892, 0.59, and 0.11.
Table \ref{tab:perlmutter_shortestnodehours} shows the parameter configurations for Perlmutter that lead to the lowest node-hours. Even with thirteen incorrect predictions, the model performs well with the $R^2$, MAE and MAPE scores of 0.95, 1.06, and 0.14.

Comparing Table \ref{tab:aurora_shortesttime} and \ref{tab:aurora_shortestnodehours} for Aurora, \ref{tab:frontier_shortesttime}
and \ref{tab:frontier_shortestnodehours} for Frontier, and \ref{tab:perlmutter_shortesttime} and \ref{tab:perlmutter_shortestnodehours}
 for Perlmutter, we see the model's behavior change for two different goals across all three machines. For the shortest-time goal, the model chooses larger node counts for most problem sizes $(O, V)$, often in the 100--500 range, to reduce wall time. In contrast, for the budget goal, the model suggests smaller node counts, in the range of 5--90, for the same problem sizes $(O, V)$ to minimize user cost.

 \begin{table}[tp]
    \centering
  \begin{minipage}{0.49\textwidth}
        \centering
        \caption{Aurora shortest time results.}
        \label{tab:aurora_shortesttime}
        \resizebox{\linewidth}{!}{
            \begin{tabular}{|r|r|r|r|r|}
            \toprule
            O & V & Nodes & Tile size & Runtime (s) \\
            \midrule
            44 & 260 & 5 & 40 & {{17.41}} \\
            81 & 835 & 185 & 80 & {{66.81}} \\
            85 & 698 & 220 & 60 & {{47.05}} \\
            99 & 718 & 260 & 60 & {{53.83}} \\
            99 & 1021 & 400 & 60 & {{112.70}} \\
            116 & 575 & 240(220) & 60 & {{38.35}}({{38.78}}) \\
            116 & 840 & 350 & 60 & {{79.95}} \\
            116 & 1184 & 400 & 80 & {{180.30}} \\
            134 & 523 & 200 & 70 & {{41.86}} \\
            134 & 951 & 400 & 70 & {{122.95}} \\
            134 & 1200 & 800 & 80 & {{196.70}} \\
            146 & 278 & 90 & 70 & {{17.92}} \\
            146 & 591 & 120 & 70(80) & {{62.89}}({{66.18}}) \\
            146 & 1096 & 300 & 73 & {{186.18}} \\
            146 & 1568 & 800(900) & 80 & {{393.72}}({{397.1}}) \\
            180 & 720 & 220 & 70 & {{104.36}} \\
            180 & 1070 & 320 & 80 & {{232.88}} \\
            196 & 764 & 300 & 80 & {{124.95}} \\
            204 & 969 & 320 & 90 & {{214.17}} \\
            235 & 1007 & 400 & 100 & {{291.99}} \\
            280 & 1040 & 110 & 100 & {{605.93}} \\
            345 & 791 & 400 & 110 & {{282.83}} \\
            \bottomrule
            \end{tabular}
        }
    \end{minipage}
\end{table}
\begin{table}[tp]
\centering
   \begin{minipage}[t]{0.49\textwidth}
        \centering
        \caption{Frontier shortest time results.}
        \label{tab:frontier_shortesttime}
        \resizebox{\linewidth}{!}{
            \begin{tabular}{|r|r|r|r|r|}
            \toprule
            O & V & Nodes & Tile size & Runtime (s)\\
            \midrule
            49 & 663 & 80 & 60 & {{22.24}} \\
            81 & 835 & 185 & 80 & {{50.86}} \\
            85 & 698 & 75 & 90 & {{56.81}} \\
            99 & 718 & 200 & 80 & {{42.24}} \\
            99 & 1021 & 200 & 80 & {{108.58}} \\
            116 & 575 & 200 & 70 & {{28.54}} \\
            116 & 840 & 300(220) & 70(90) & {{59.17(63.07)}} \\
            116 & 1184 & 350 & 70 & {{159.66}} \\
            134 & 523 & 350(300) & 70(80) & {{25.95(28.9)}} \\
            134 & 951 & 400(300) & 70(90) & {{90.15(96.44)}} \\
            134 & 1200 & 700 & 80 & {{135.81}} \\
            146 & 591 & 70(120) & 90(120) & {{57.82(60.36)}} \\
            146 & 1096 & 700 & 90(73) & {{108.84(116.43)}} \\
            180 & 720 & 350 & 80 & {{71.43}} \\
            180 & 1070 & 400 & 90 & {{172.50}} \\
            196 & 764 & 350 & 90 & {{91.41}} \\
            204 & 969 & 350 & 100 & {{158.46}} \\
            235 & 1007 & 400 & 100 & {{222.96}} \\
            280 & 1040 & 600 & 100 & {{249.16}} \\
            345 & 791 & 350 & 130 & {{249.66}} \\
            \bottomrule
            \end{tabular}
        }
    \end{minipage}
\end{table}
\begin{table}[tp]
    \centering
    \begin{minipage}[t]{0.49\textwidth}
        \centering
        \caption{Perlmutter shortest time results.}
        \label{tab:perlmutter_shortesttime}
        \resizebox{\linewidth}{!}{
            \begin{tabular}{|r|r|r|r|r|}
            \toprule
            O & V & Nodes & Tile size & Runtime (s)\\
            \midrule
            49 & 663 & 150(60) & 70 & {{19.28(25.72)}} \\
            81 & 835 & 185(240) & 90(80) & {{64.26(71.88)}} \\
            85 & 698 & 350 & 80 & {{28.29}} \\
            99 & 718 & 260(400) & 80 & {{41.87(44.18)}} \\
            99 & 1021 & 260 & 90 & {{120.94}} \\
            116 & 575 & 300 & 80(90) & {{24.89(28.6)}} \\
            116 & 840 & 350(400) & 70(80) & {{65.78(76.38)}} \\
            116 & 1184 & 350 & 80 & {{218.27}} \\
            134 & 523 & 300(260) & 70(110) & {{24.47(37.04)}} \\
            134 & 951 & 350(400) & 80(90) & {{98.11(144.71)}} \\
            134 & 1200 & 300(260) & 100(120) & {{190.06((213.59))}} \\
            146 & 278 & 110 & 70 & {{10.48}} \\
            146 & 591 & 110(150) & 100(120) & {{59.23(60.0)}} \\
            146 & 1096 & 400(500) & 120(90) & {{140.22(142.3)}} \\
            180 & 720 & 280 & 90 & {{95.97}} \\
            180 & 1070 & 350 & 90 & {{216.20}} \\
            196 & 764 & 350(300) & 100 & {{122.36(122.8)}} \\
            204 & 969 & 350(260) & 120 & {{190.32(229.83)}} \\
            235 & 1007 & 400 & 130(110) & {{340.19(353.89)}} \\
            280 & 1040 & 400(500) & 120 & {{323.81(337.68)}} \\
            345 & 791 & 260(400) & 130(140) & {{282.05(292.52)}} \\
            \bottomrule
            \end{tabular}
        }
   \end{minipage}
\end{table}
\begin{table}[tp]
    \centering
    \caption{Aurora shortest node hours results.}
    \label{tab:aurora_shortestnodehours}
\begin{tabular}{|r|r|r|r|r|r|}
\toprule
Oa & Va & Nodes & tilesize & Runtime(s) & Node Hours \\
\midrule
44 & 260 & 5 & 40 & {{17.41}} & {{0.02}} \\
81 & 835 & 25 & 80 & {{193.26}} & {{1.34}} \\
85 & 698 & 15 & 120 & {{146.45}} & {{0.61}} \\
99 & 718 & 15 & 110(90) & {{173.41}}(182.32) & {{0.72}}(0.76) \\
99 & 1021 & 35 & 110 & {{285.94}} & {{2.78}} \\
116 & 575 & 15 & 90 & {{123.51}} & {{0.51}} \\
116 & 840 & 35 & 90 & {{178.26}} & {{1.73}} \\
116 & 1184 & 15 & 120(140) & {{682.15}}(706.92) & {{2.84}}(2.95)\\
134 & 523 & 65 & 90 & {{58.25}} & {{1.05}} \\
134 & 951 & 35(25) & 130(140) & {{282.70}}(565.37) & {{2.75}}(3.93) \\
134 & 1200 & 45 & 120 & {{469.57}} & {{5.87}} \\
146 & 278 & 10 & 120(100) & {{38.67}}(38.83) & {{0.11}}(0.11) \\
146 & 591 & 30 & 100 & {{102.96}} & {{0.86}} \\
146 & 1096 & 30 & 140 & {{498.74}} & {{4.16}} \\
146 & 1568 & 200 & 90 & {{616.39}} & {{34.24}} \\
180 & 720 & 20 & 130 & {{293.36}} & {{1.63}} \\
180 & 1070 & 30 & 120 & {{591.97}} & {{4.93}} \\
196 & 764 & 50 & 110 & {{247.22}} & {{3.43}} \\
204 & 969 & 90 & 90 & {{380.81}} & {{9.52}} \\
235 & 1007 & 25 & 140 & {{907.16}} & {{6.30}} \\
280 & 1040 & 50(40) & 130(140) & {{876.74}}(1163.77) & {{12.18}}(12.93) \\
345 & 791 & 50 & 130 & {{589.65}} & {{8.19}} \\
\bottomrule
\end{tabular}
\end{table}
\begin{table}[tp]
    \centering
    \caption{Frontier shortest node hours results.}
    \label{tab:frontier_shortestnodehours}
\begin{tabular}{|r|r|r|r|r|r|}
\toprule
Oa & Va & Nodes & tilesize & Runtime(s) & Node Hours \\
\midrule
49 & 663 & 10 & 150 & {{64.67}} & {{0.18}} \\
81 & 835 & 15 & 150(110) & {{157.31}}(162.3) & {{0.66}}(0.68)\\
85 & 698 & 15 & 110(140) & {{104.37}}(114.63) & {{0.43}}(0.48) \\
99 & 718 & 25 & 130 & {{114.69}} & {{0.80}} \\
99 & 1021 & 35 & 110 & {{206.02}} & {{2.00}} \\
116 & 575 & 15 & 130(180) & {{116.08}}(143.36) & {{0.48}} (0.6)\\
116 & 840 & 35 & 160(150) & {{211.50}}(216.3) & {{2.06}}(2.10) \\
116 & 1184 & 65 & 130 & {{300.80}} & {{5.43}} \\
134 & 523 & 15 & 120 & {{104.93}} & {{0.44}} \\
134 & 951 & 95(35) & 130(160) & {{154.45}}(425.8) & {{4.08}}(4.14) \\
134 & 1200 & 45 & 140 & {{467.01}} & {{5.84}} \\
146 & 591 & 30 & 120 & {{98.69}} & {{0.82}} \\
146 & 1096 & 30(50) & 150(140) & {{532.34}}(426.38) & {{4.44}}(5.92) \\
180 & 720 & 50 & 90 & {{172.48}} & {{2.40}} \\
180 & 1070 & 90 & 150 & {{313.89}} & {{7.85}} \\
196 & 764 & 50 & 150(110) & {{203.36}}(205.37) & {{2.82}}(2.85) \\
204 & 969 & 70(50) & 140(120) & {{323.80}}(463.97) & {{6.30}}(6.44) \\
235 & 1007 & 50 & 150 & {{563.71}} & {{7.83}} \\
280 & 1040 & 70(50) & 140(130) & {{666.34}}(1175.96) & {{12.96}}(16.33) \\
345 & 791 & 50 & 150 & {{606.59}} & {{8.42}} \\
\bottomrule
\end{tabular}
\end{table}
\begin{table}[tp]
    \centering
    \caption{Perlmutter shortest node hours results.}
    \label{tab:perlmutter_shortestnodehours}
\begin{tabular}{|r|r|r|r|r|r|}
\toprule
Oa & Va & Nodes & tilesize & Runtime(s) & Node Hours \\
\midrule
49 & 663 & 30(60) & 120(70) & {{ 41.93(25.72)}} & {{0.35(0.43)}} \\
81 & 835 & 25 & 150(160) & {{137.98(138.24)}} & {{0.96(0.96)}}\\
85 & 698 & 25 & 110(130)  & {{84.25}}(111.05	) & {{0.59}}(0.77) \\
99 & 718 & 25 & 110(130) & {{110.90(125.04)}} & {{0.77(0.86)}} \\
99 & 1021 & 35 & 110 & {{247.86}} & {{2.41}} \\
116 & 575 & 45(25) & 110(60) & {{ 72.77(152.16)}} & {{0.91(1.06)}}\\
116 & 840 & 55 & 130() & {{167.37(195.06)}} & {{2.56(2.98)}} \\
116 & 1184 & 75 & 140 & {{381.18}} & {{7.94}} \\
134 & 523 & 35 & 130 & {{67.11}} & {{0.65}} \\
134 & 951 & 55 & 110(100) & {{342.45}}(372.17) & {{5.23}}(5.69) \\
134 & 1200 & 150(10) & 130(90) & {{292.53(487.83)}} & {{12.19(15.58)}} \\
146 & 591 & 50(10) & 120(90) & {{78.11(421.38)}} & {{1.08(1.17)}} \\
146 & 1096 & 60 & 130 & {{512.14}} & {{8.54}} \\
180 & 720 & 50 & 110 & {{210.81}} & {{2.93}} \\
180 & 1070 & 130 & 120 & {{326.66}} & {{11.80}} \\
196 & 764 & 70 & 140 & {{227.66(242.44)}} & {{4.43}}(4.71) \\
204 & 969 & 110 & 150(140) & {{311.10(344.44)}}& {{9.51(10.52)}} \\
235 & 1007 & 115 & 110 & {{518.10}} & {{16.55}} \\
280 & 1040 & 130(180) & 110(120) & {{926.60}}(797.0) & {{33.46}}(39.85) \\
345 & 791 & 220(150) & 120 & {{297.69(466.52)}} & {{18.19(19.43)}} \\
\bottomrule
\end{tabular}
\end{table}

\subsection{Active learning results}

\begin{figure*}[tp]
  \centering
    \includegraphics[scale=0.4]{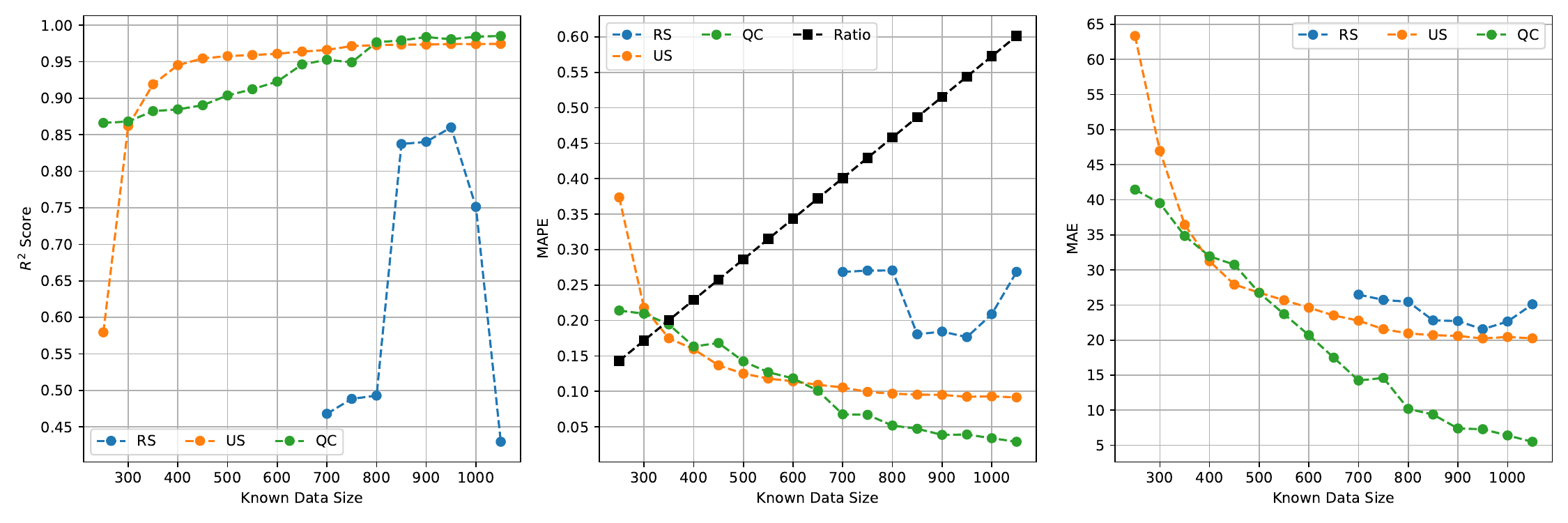}
    \vspace*{-20pt}
    \caption{Aurora active learning results.}
    \label{fig:auroraactivelearning}
\end{figure*}

\begin{figure*}[tp]
  \centering
  \includegraphics[scale=0.4]{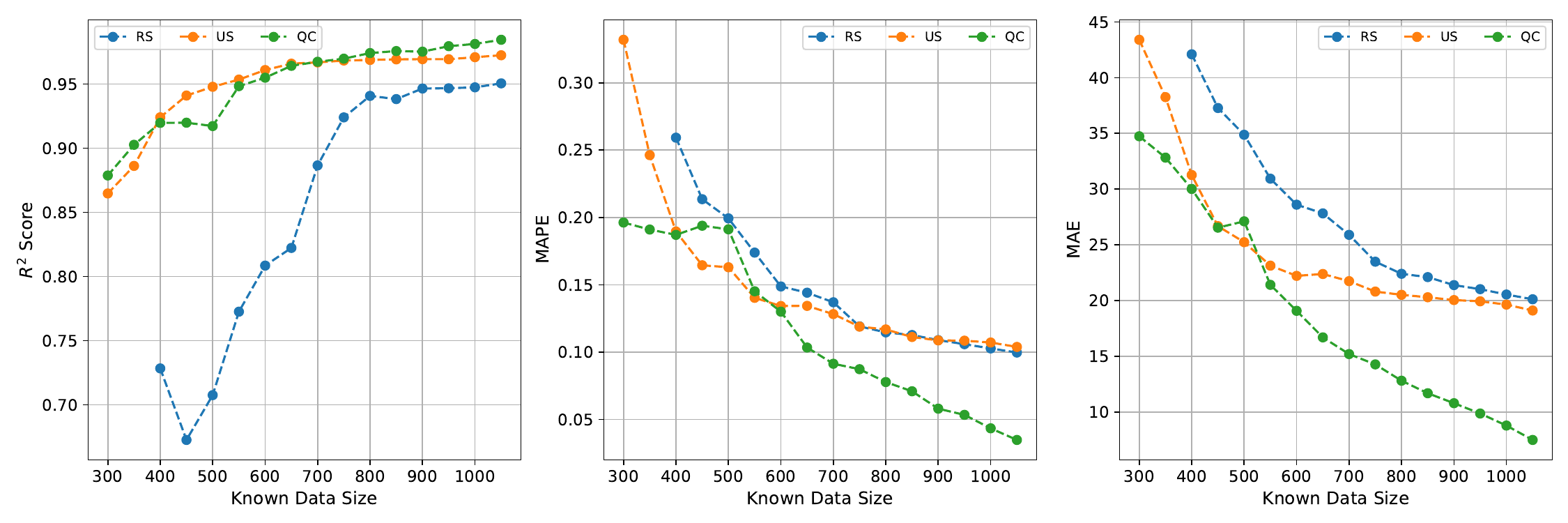}
  \vspace*{-20pt}
    \caption{Frontier active learning results.}
    \label{fig:frontieractivelearning}
\end{figure*}

\begin{figure*}[tp]
  \centering
  \includegraphics[scale=0.4]{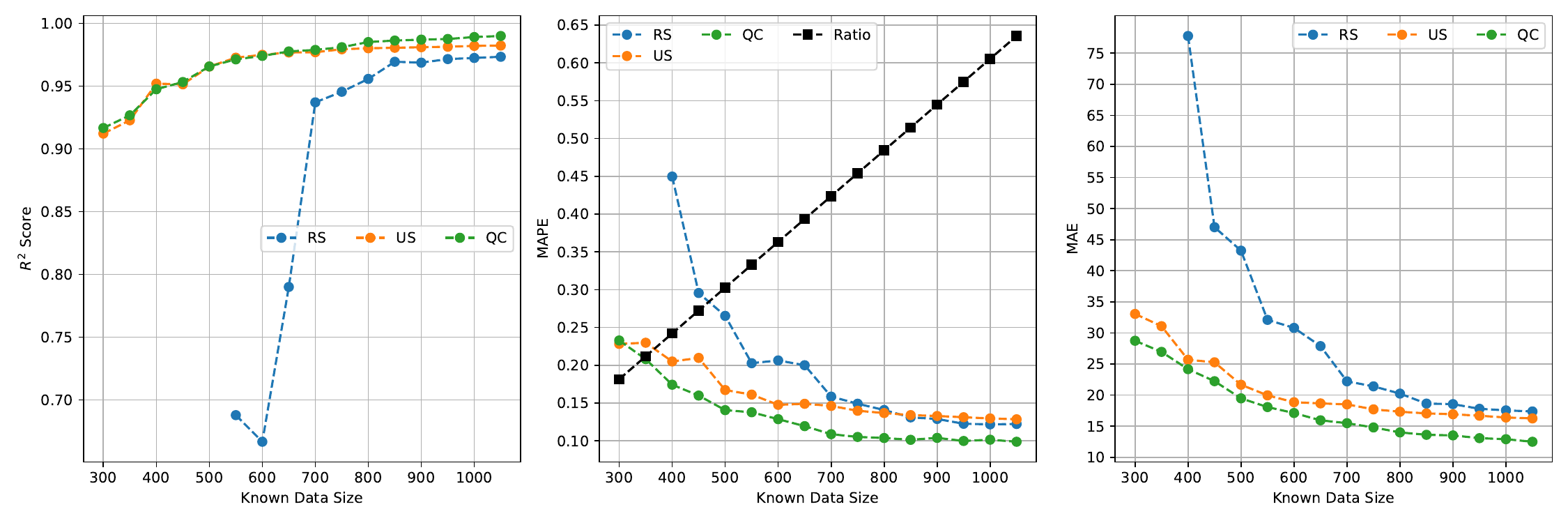}
    \vspace{-20pt}
    \caption{Perlmutter active learning results.}
    \label{fig:perlmutteractivelearning}
\end{figure*}

The performance evolution of the active-learning models for Aurora, Frontier, and Perlmutter is illustrated in Figures~\ref{fig:auroraactivelearning}, \ref{fig:frontieractivelearning}, and \ref{fig:perlmutteractivelearning}. The y-axes show the evaluation metrics $R^2$ score, MAPE, and MAE, while the x-axes represent the number of data instances obtained from runs on the supercomputers. We consider three query strategies: Random Sampling (RS), which serves as the baseline, Uncertainty Sampling (US), and Query-by-Committee (QC). Predicting optimal parameter configurations from predicted execution times is challenging because incomplete information introduces performance variance. Therefore, the active-learning experiments evaluate both the accuracy of execution-time prediction and the ability to determine the corresponding parameter configurations for the STQ.

For the non-STQ objective, the active-learning results show different behavior across the three platforms. In Figure~\ref{fig:auroraactivelearning}, the baseline RS performs poorly on Aurora. Results are shown only once there are 700 or more known data points because below that, $R^2$ scores are negative and both MAPE and MAE exceed the plot range. Figure~\ref{fig:frontieractivelearning} presents the active-learning results on Frontier for the non-STQ objective. Here, the performance of baseline RS is better and more consistent than it is on Aurora. The non-STQ active-learning results for Perlmutter are shown in Figure~\ref{fig:perlmutteractivelearning}. For Perlmutter, the baseline RS performs well after 700 data points.


\subsubsection*{Shortest-time question}

For Aurora, Figure~\ref{fig:aurorastqbq} shows that although the STQ goal increases variance in performance, models trained for STQ outperform their non-STQ models. Furthermore, US achieves better results than QC. The main finding is that with only 450 data points, which is 25\% of the original dataset, a MAPE of 0.2 is achieved. The MAPE is further improved to 0.1 with approximately 550 experiments. In both cases, the models maintain high $R^2$ scores of around 0.98.

For Frontier, Figure~\ref{fig:frontierstqbq} presents the active-learning results for the STQ objective. RS under STQ fails on Frontier as it does on Aurora, and US outperforms QC overall. Furthermore, Frontier proves more difficult to predict than Aurora. The performance variance introduced by the STQ goal is noticeable as expected. For Frontier under the STQ objective, a MAPE of about 0.20 is achievable with 450 to 650 experiments, which corresponds to 20\%--30\% of the original dataset, and a value near 0.10 is achievable with around 850 experiments, which corresponds to 47\% of the original dataset.

For Perlmutter, Figure~\ref{fig:perlmutterstqbq} shows the active-learning results for the STQ objective. Under the STQ goal, RS performs poorly. Unlike Aurora and Frontier, QC outperforms US for Perlmutter. The performance variance caused by the STQ goal is less prominent compared to Aurora or Frontier. For Perlmutter, a MAPE of about 0.2 is achieved with 500 to 750 experiments, which corresponds to 25\%--35\% of the original dataset, and by 900 data points, the MAPE is improved to about 0.1.

\subsubsection*{Budget question}
\begin{figure*}[tp]
    \centering
    \includegraphics[scale=0.4]{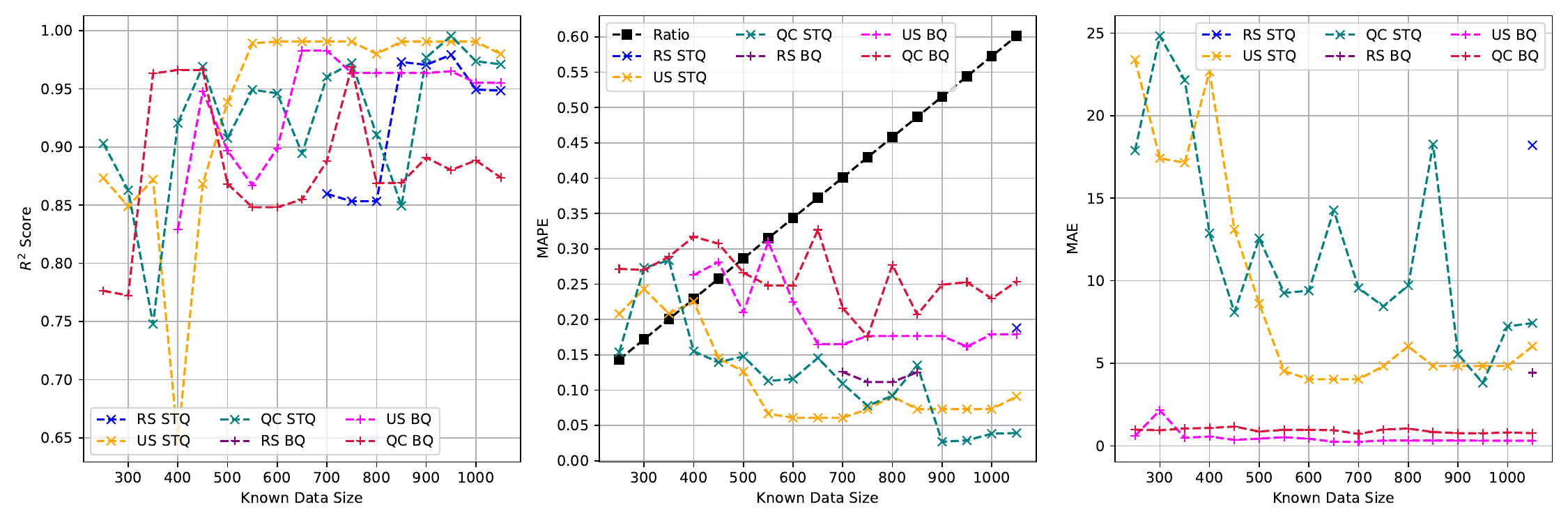}
    \vspace{-20pt}
    \caption{Aurora active learning results for shortest time and budget question.}
    \label{fig:aurorastqbq}
\end{figure*}

\begin{figure*}[tp]
    \centering
    \includegraphics[scale=0.4]{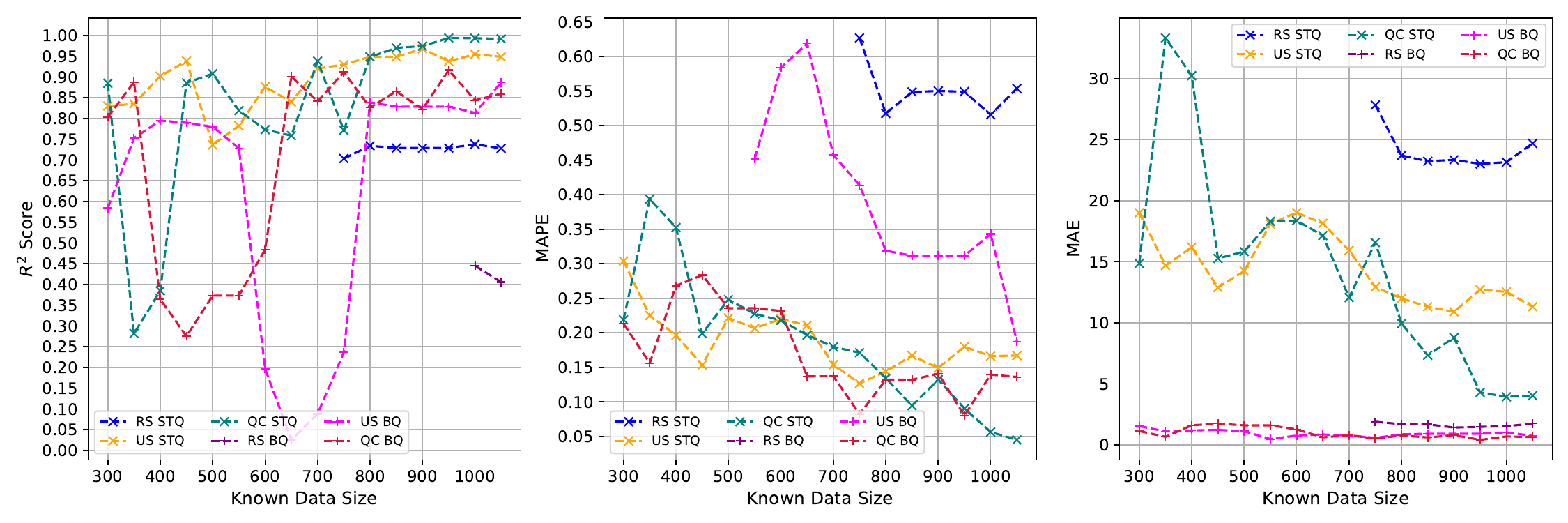}
    \vspace{-20pt}
    \caption{Frontier active learning results for shortest time and budget question.}
    \label{fig:frontierstqbq}
\end{figure*}

\begin{figure*}[tp]
    \centering
    \includegraphics[scale=0.4]{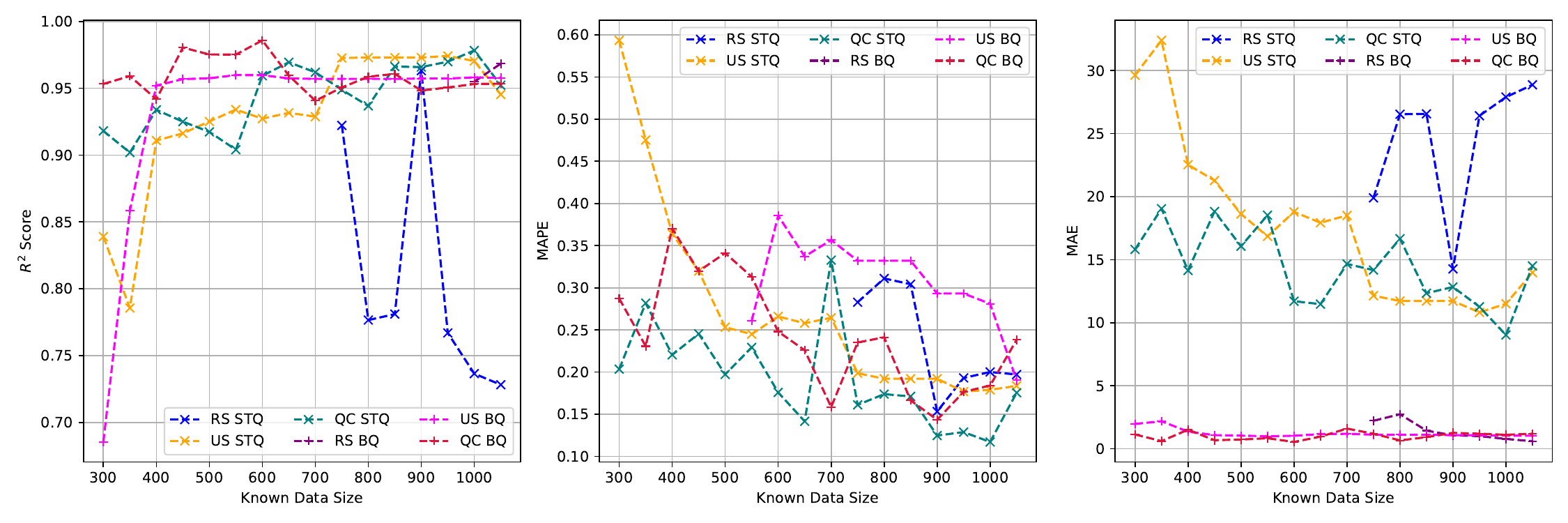}
    \vspace{-20pt}
    \caption{Perlmutter active learning results for shortest time and budget question.}
    \label{fig:perlmutterstqbq}
\end{figure*}

Figures~\ref{fig:aurorastqbq}, \ref{fig:frontierstqbq}, and \ref{fig:perlmutterstqbq} show the active-learning results for the budget question. 
For the BQ goal in Aurora, we can see that US outperforms QC similar to the STQ goal. It reaches a MAPE score of about 0.17 with 650 experiments. Unlike the STQ goal in Frontier, QC performs better than US for the BQ goal, achieving a MAPE of less than 0.15 with 650 experiments onwards. It reaches values close to 0.10 at larger known data sizes. For the BQ goal in Perlmutter, QC achieves better results compared to US, and results in a MAPE of 0.25 with 650 experiments. These results indicate that the BQ behavior differs from the STQ case: US is better on Aurora, while QC is better on Frontier and Perlmutter.


\subsection{Generative learning results}
We evaluate Gaussian Copula and CTGAN synthesizers as introduced earlier.
First, investigating the overheads of data generation, Table \ref{tab:generationtimes} show the fitting and data generation (sampling) 
times for the two synthesizers with the corresponding standard deviations. Since the results are very similar for Aurora, Frontier, and Perlmutter we only report those for Frontier. CTGAN based data fitting is about 160$\times$ more expensive than Gaussian Copula based data fitting.
On the other hand, data generation with CTGAN takes about the same time as Gaussian Copula.

\begin{table}[tp]
    \centering
    \caption{Fitting and sampling times for the synthesizers.}
    \label{tab:generationtimes}
    \begin{tabular}{|c|c|c|} \hline
         & Gaussian Copula & CTGAN \\ \hline
      Fitting to train dataset & 515 ms ± 79.3 ms & 1min 23s ± 14.2 s \\ \hline
      Sampling 100 samples & 26.3 ms ± 3.36 ms  & 25.3 ms ± 2.35 ms \\ \hline
      Sampling 500 samples &  33.2 ms ± 1.96 ms & 34.7 ms ± 10 ms \\ \hline
      Sampling 1000 samples & 39.7 ms ± 11.5 ms  & 42.8 ms ± 12.5 ms \\ \hline
    \end{tabular}
\end{table}

\subsubsection*{Shortest-time question}

\begin{figure*}[tp]
  \centering
  \includegraphics[scale=0.4]{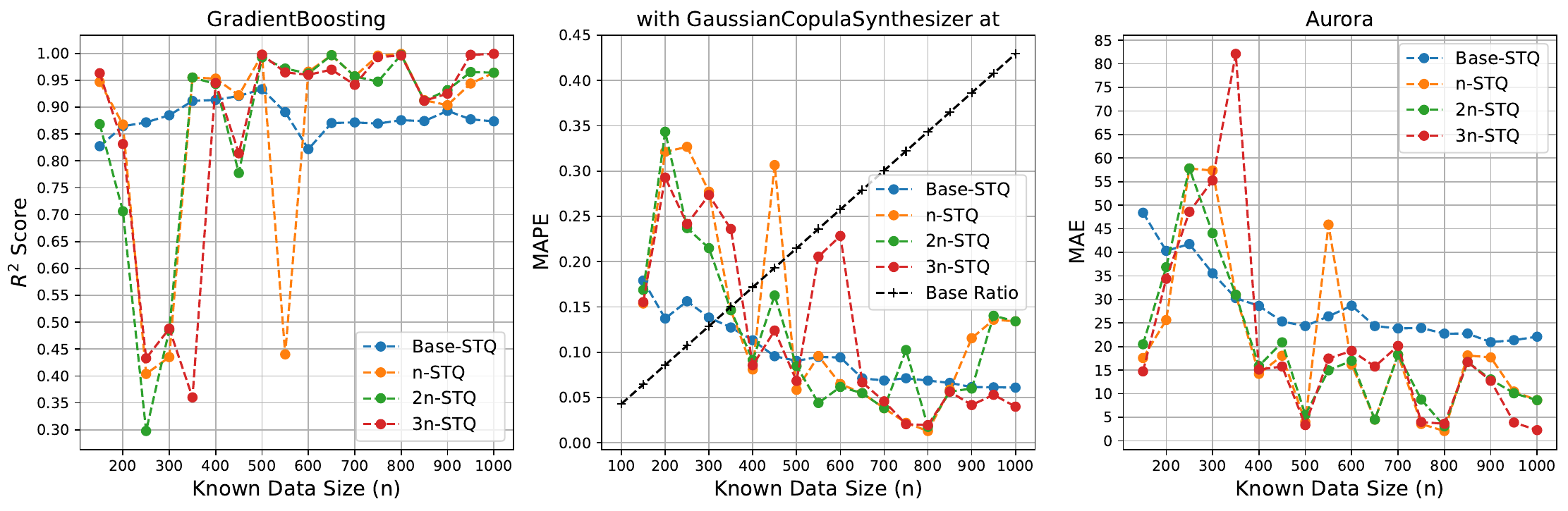}
    \caption{Aurora generative learning results for shortest-time question --- Gaussiancopula}
    \label{fig:auroragaussiancopula}
\end{figure*}

\begin{figure*}[tp]
  \centering
  \includegraphics[scale=0.4]{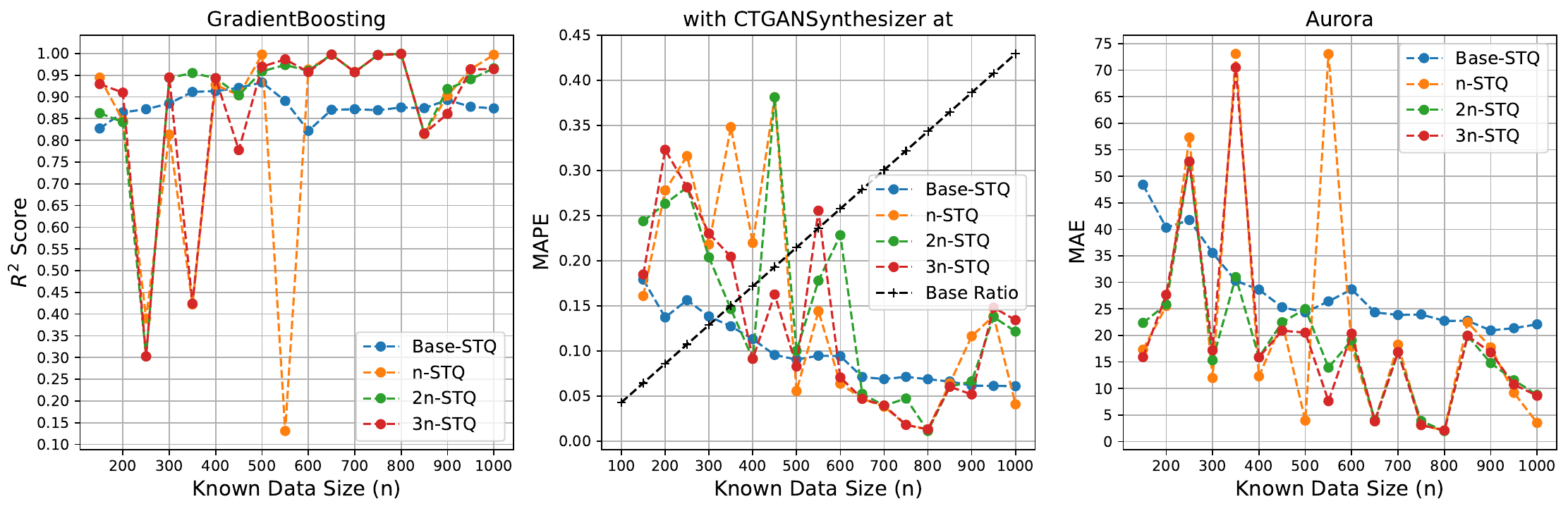}
    \caption{Aurora generative learning results for shortest-time question --- CTGAN}
    \label{fig:auroractgan}
\end{figure*}

Figures \ref{fig:auroragaussiancopula} and \ref{fig:auroractgan} show the generative
results for Aurora with Gaussian Copula and CTGAN synthesizers for the STQ goal. It also shows the impact of the number of synthetic data points on the performance of an ML model by generating 1, 2, and 3$\times$ of the number of the known data points, separately.
From the figures, we see that synthetic data causes instability in learning.
Compared with active learning, generative learning is modestly outperforming, especially in terms of MAPE. With Gaussian Copula,
a MAPE of about and less than 0.07 is achieved with 500 and more experiments, i.e. data points, with the generation of 3$\times$ the number of known (base) data points. Impressively, with 600 or more experiments and the generation of 3$\times$ the number of known data points, a MAPE of about and less than 0.05 is achieved. However, active learning might be preferable to generative learning due to the overhead of data generation. 

\begin{figure*}[tp]
  \centering
  \includegraphics[scale=0.4]{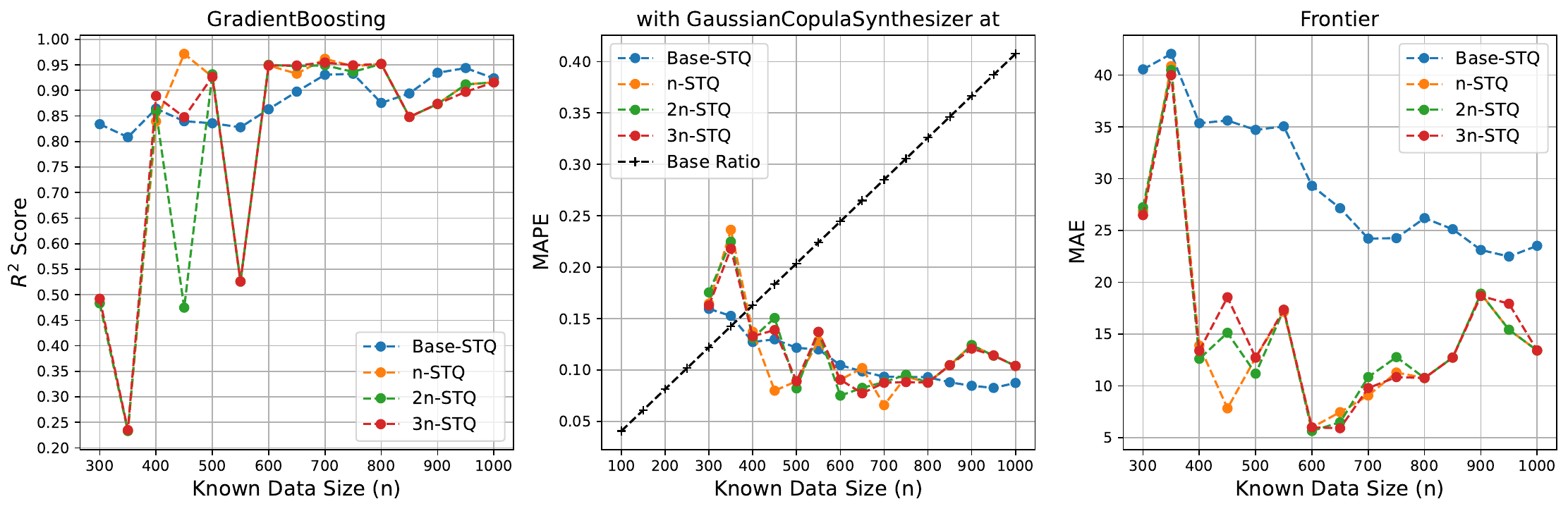}
    \caption{Frontier generative learning results for shortest-time question --- Gaussiancopula}
    \label{fig:frontiergaussiancopula}
\end{figure*}

\begin{figure*}[tp]
  \centering
  \includegraphics[scale=0.4]{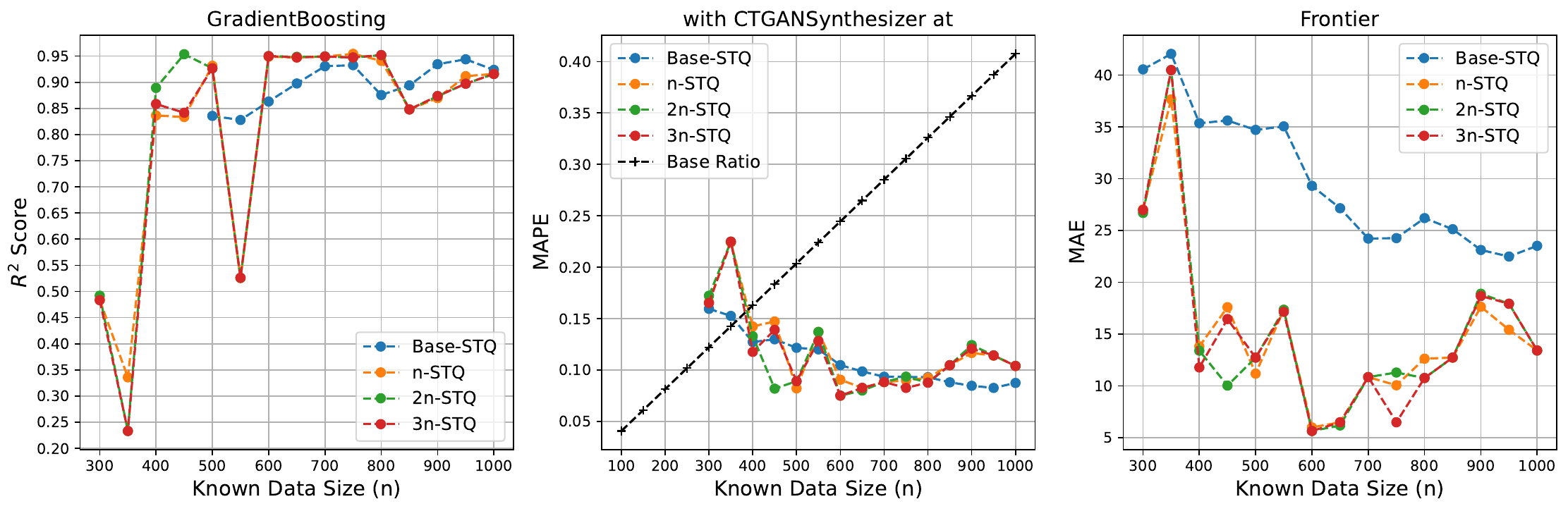}
    \caption{Frontier generative learning results for shortest-time question --- CTGAN}
    \label{fig:frontierctgan}
\end{figure*}

Figures \ref{fig:frontiergaussiancopula} and \ref{fig:frontierctgan} show the generative
results for Frontier. For both Gaussian Copula and CTGAN, a MAPE of about 0.10 is achieved with 500 or more experiments with all levels (amounts) of data generation. In contrast to Aurora, the amount of data generation does not impact the performance of the learning algorithm.

\begin{figure*}[tp]
  \centering
  \includegraphics[scale=0.4]{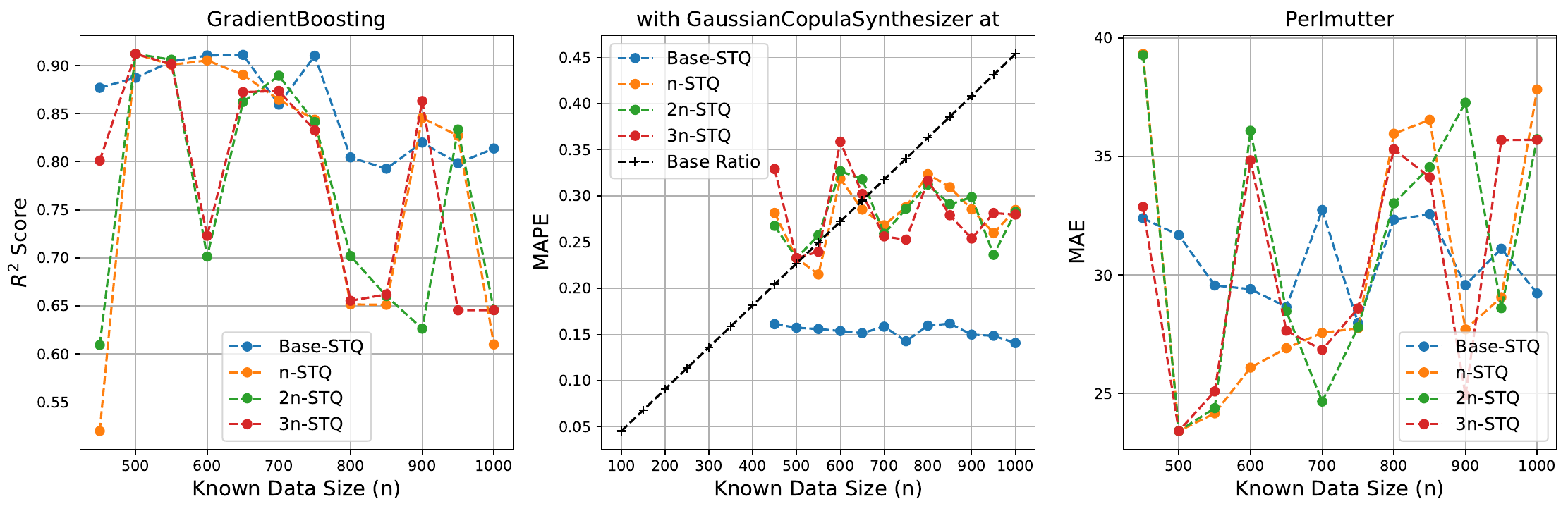}
    \caption{Perlmutter generative learning results for shortest-time question --- Gaussiancopula}
    \label{fig:perlmuttergaussiancopula}
\end{figure*}

\begin{figure*}[tp]
  \centering
  \includegraphics[scale=0.4]{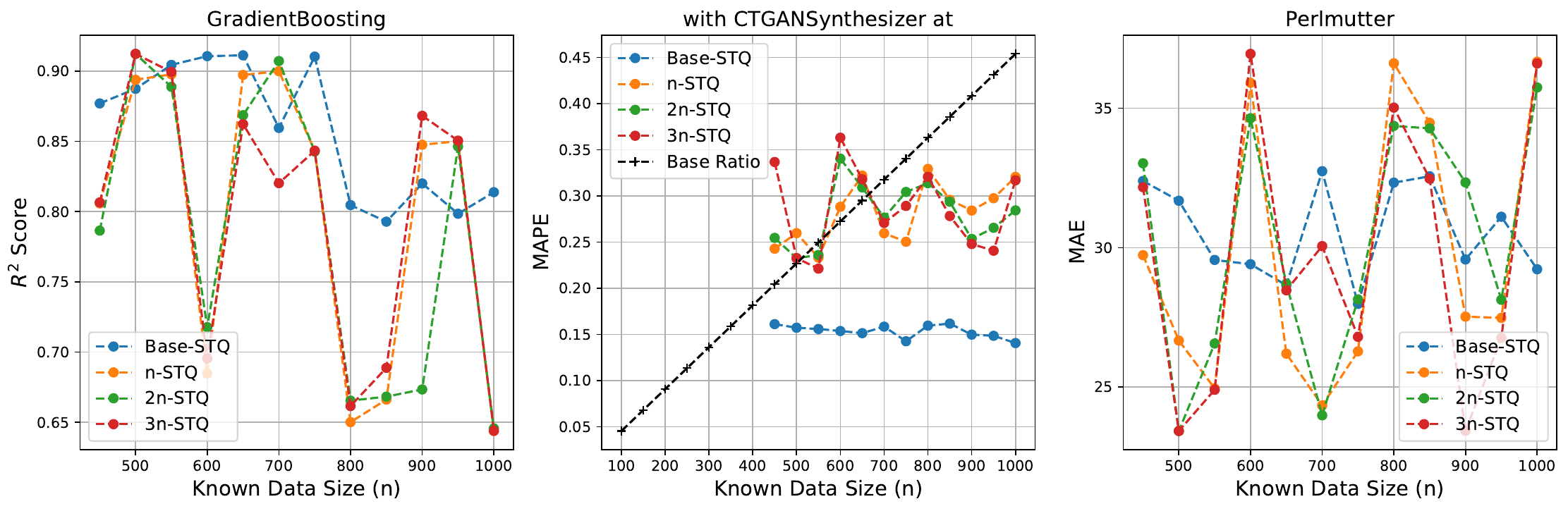}
    \caption{Perlmutter generative learning results for shortest-time question --- CTGAN}
    \label{fig:perlmutterctgan}
\end{figure*}

Figures \ref{fig:perlmuttergaussiancopula} and \ref{fig:perlmutterctgan} the generative learning results for Perlmutter with Gaussian Copula and CTGAN synthesizers for the STQ goal.
Unlike Aurora and Frontier, adding synthetic data leads to higher error for Perlmutter for both Gaussian Copula and CTGAN.




\subsubsection*{Budget question}
\begin{figure*}[tp]
  \centering
  \includegraphics[scale=0.4]{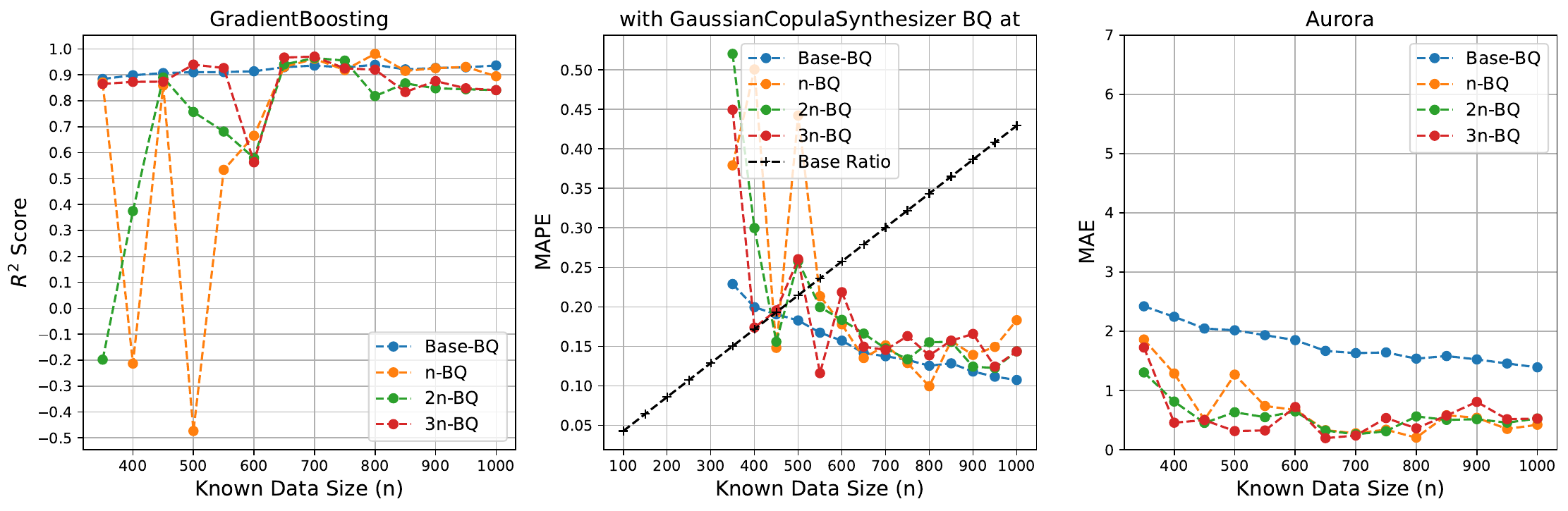}
    \caption{Aurora generative learning results for budget question --- Gaussiancopula}
    \label{fig:auroragaussiancopulabq}
\end{figure*}

\begin{figure*}[tp]
  \centering
  \includegraphics[scale=0.4]{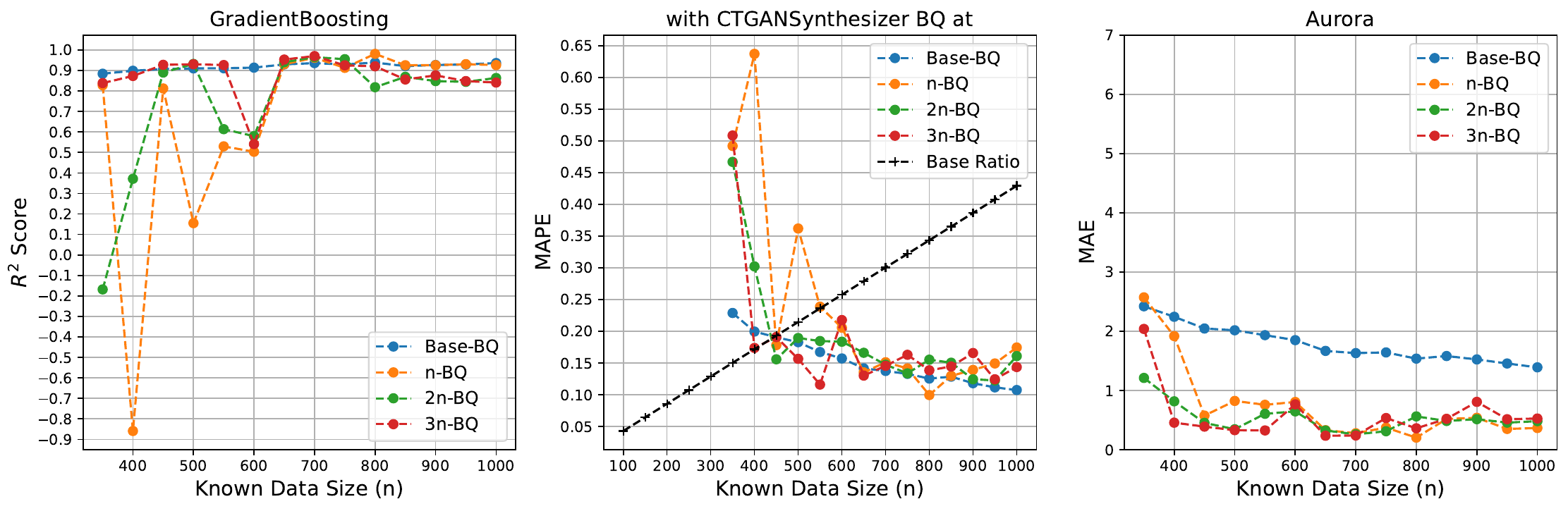}
    \caption{Aurora generative learning results for budget question --- CTGAN}
    \label{fig:auroractganbq}
\end{figure*}

The generative
results for Aurora with Gaussian Copula and CTGAN synthesizers for the BQ goal are shown in Figures \ref{fig:auroragaussiancopulabq} and \ref{fig:auroractganbq}. Similar to the STQ case, generated data can introduce instability for smaller n. However, after 500 known data points, a mape of less than 0.2 is achieved with the generation of 1, 2, and 3$\times$ of known data points.

\begin{figure*}[tp]
  \centering
  \includegraphics[scale=0.4]{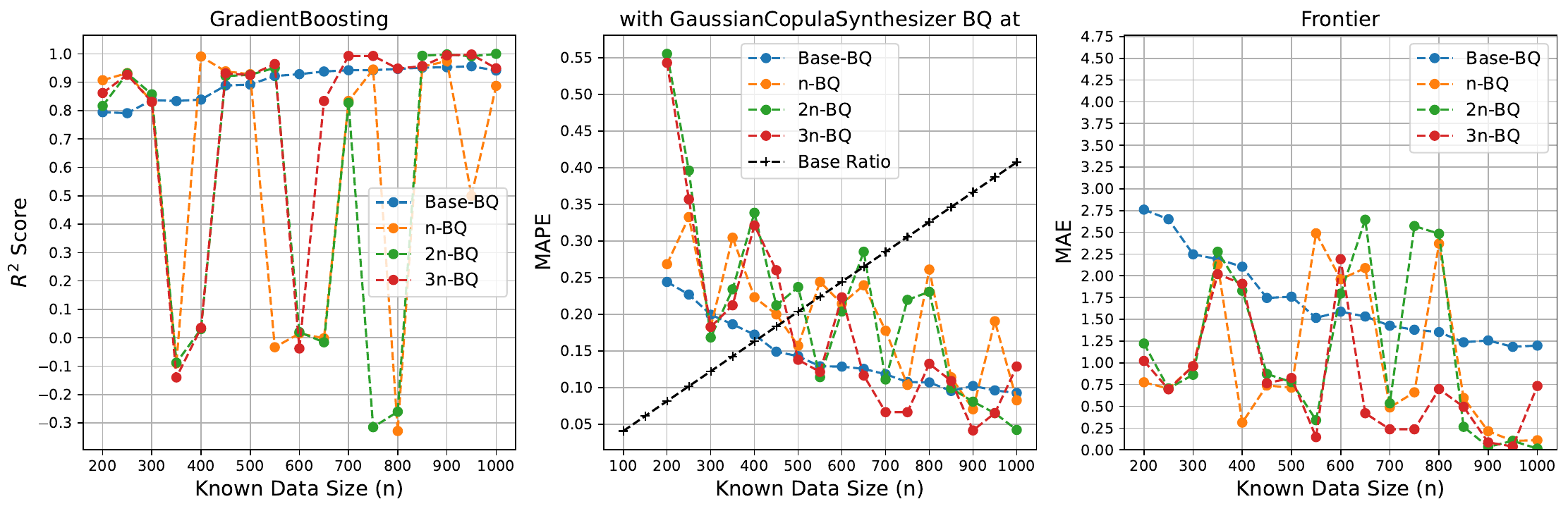}
    \caption{Frontier generative learning results for budget question --- Gaussiancopula}
    \label{fig:frontiergaussiancopulabq}
\end{figure*}

\begin{figure*}[tp]
  \centering
  \includegraphics[scale=0.4]{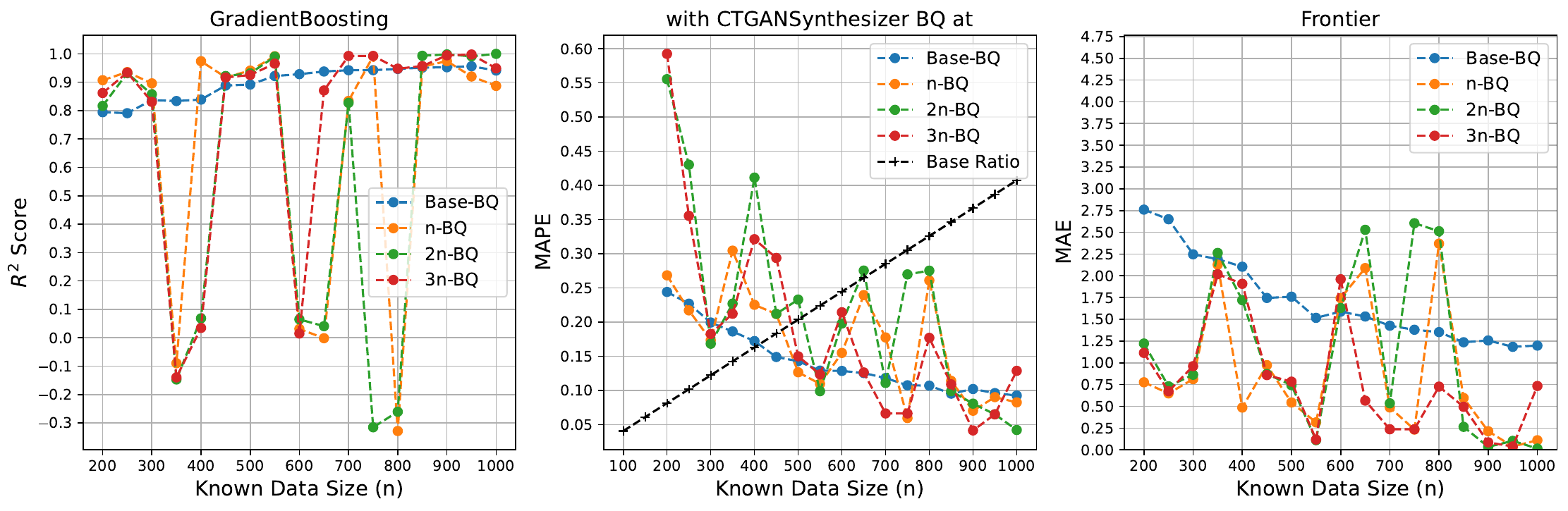}
    \caption{Frontier generative learning results for budget question -- CTGAN}
    \label{fig:frontierctganbq}
\end{figure*}

Figures~\ref{fig:frontiergaussiancopulabq} and~\ref{fig:frontierctganbq} show the generative
results for Frontier for Gaussian Copula and CTGAN synthesizers. Compared to Aurora, Frontier shows greater inconsistency in learning throughout the entire range of training sizes.

\begin{figure*}[tp]
  \centering
  \includegraphics[scale=0.4]{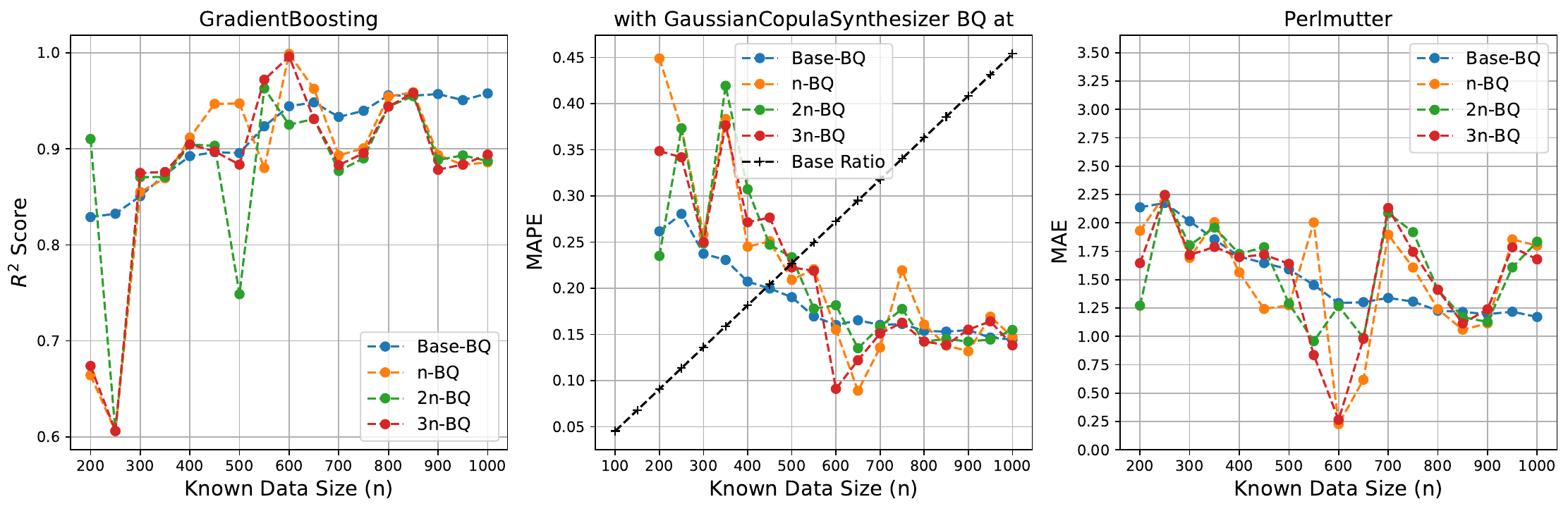}
    \caption{Perlmutter generative learning results for budget question --- Gaussiancopula}
    \label{fig:perlmuttergaussiancopulabq}
\end{figure*}

\begin{figure*}[tp]
  \centering
  \includegraphics[scale=0.4]{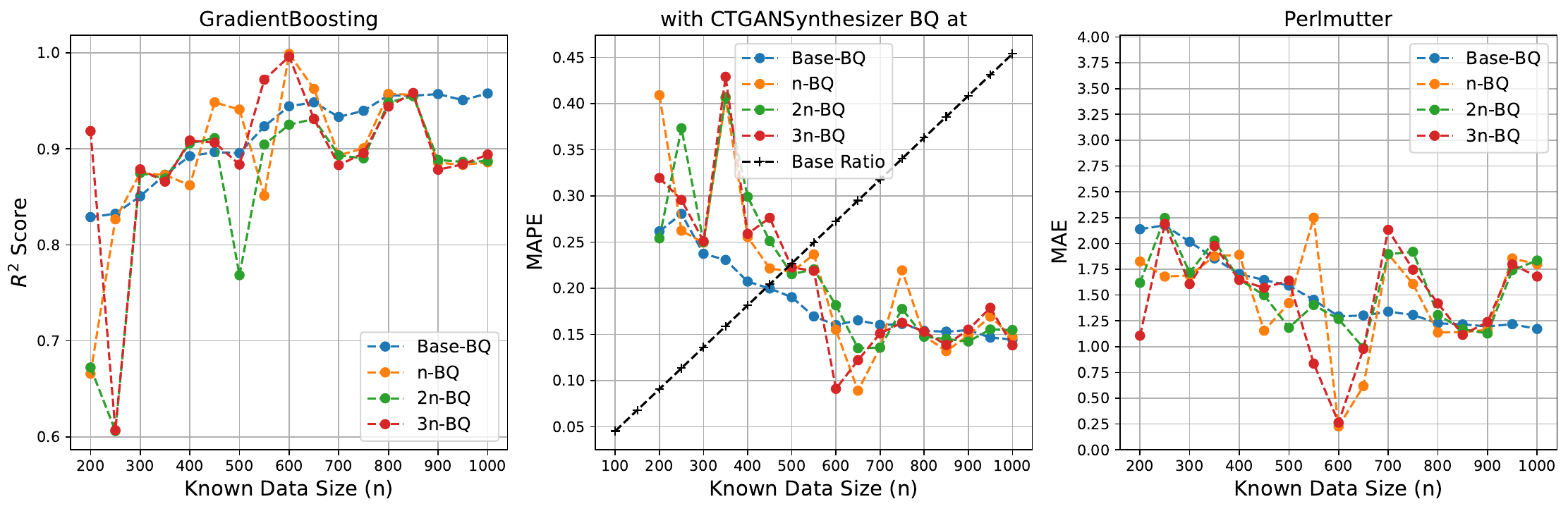}
    \caption{Perlmutter generative learning results for budget question --- CTGAN}
    \label{fig:perlmutterctganbq}
\end{figure*}

Figures \ref{fig:perlmuttergaussiancopulabq} and \ref{fig:perlmutterctganbq} present the generative-learning results for Perlmutter under the BQ objective using Gaussian Copula and CTGAN\@. For Perlmutter, a mape of about 0.15 is achieved with 600 or more data points.






\subsection{Statistical analysis for runtime}

To evaluate the computational characteristics of batches selected by each active-learning strategy or generated by our generative learning methods, we perform a statistical analysis of the item runtimes at each iteration, where each runtime corresponds to a unique combination of problem size and configuration. For a given iteration $t \in \{1, \dots, T\}$, where a batch of $n$ items is selected, let $r_i^{(t)}$ denote the runtime in seconds of the $i$-th item in the batch. We employ three key statistical metrics to characterize the set of runtimes $\{r_i^{(t)}\}_{i=1}^{n}$ for each iteration.

The mean runtime, $\bar{r}^{(t)}$, is calculated as the arithmetic average of the runtimes within the batch selected or generated at iteration $t$:
\[
\bar{r}^{(t)} \;=\; \frac{1}{n}\sum_{i=1}^{n} r_i^{(t)}
\]

We use the variance $v_r^{(t)}$ in seconds squared to measure the dispersion in experiment runtimes within the batch at each iteration, which is defined as follows:\[
v_{r}^{(t)} \;=\; \frac{1}{n}\sum_{i=1}^{n}\!\left(r_i^{(t)} - \bar{r}^{(t)}\right)^{2}
\]

The standard deviation \(s_r^{(t)}\) is the root mean square distance (RMS) of the experiment run times from the batch mean at iteration \(t\). It is linked to the variance by
\[
s_r^{(t)} \;=\; \sqrt{\,v_r^{(t)}\,}
\quad\text{(seconds)}.
\]

\subsection{Timing analysis of selected samples}
We analyze 20 active-learning iterations. Each iteration consists of a batch of 50 experiments. For each batch, we compute the mean runtime, standard deviation, and variance. In Figures \ref{fig:aurorastats}, \ref{fig:frontierstats}, and \ref{fig:perlmutterstats}, we plot these statistics on the y-axis across iterations \(t\) on the x-axis for the three active learning strategies for Aurora, Frontier, and Perlmutter.

A higher mean runtime indicates that the active learning algorithm is selecting more computationally expensive experiments in that iteration, while lower values indicate low computational cost. We plot $\bar{r}^{(t)}$ to observe whether active learning strategies select faster or slower batches across iterations.
The variance measures the heterogeneity of runtimes within a selected batch. A large variance means that the batch is composed of experiments with widely different runtimes, that is, a mix of longer and shorter processing times. On the other hand, a small variance indicates that the experiments within the batch are more similar in their runtime.
Larger standard deviation indicate more variable batches, and smaller values imply that the active learning algorithm is selecting runtimes that are similar in value. We plot \(s_r^{(t)}\) across iterations for an easier interpretation of variance since it has the same unit, seconds, as the runtime.

\subsubsection*{Mean}
From Figure \ref{fig:aurorastats}, QC begins with the highest mean runtime, about 580 s, while US starts around 340 s and RS around 260--280 s on Aurora. The initially high mean runtime of the QC-selected batch indicates that the committee first disagreed most on slower configurations. After these configurations are learned, QC selects batches with faster configurations. US also selects batches with larger runtimes early, but after that it stays mostly within a stable range of about 200--380 s. Compared to QC and US, the baseline RS stays close to average runtime because it does not use model feedback.

We see from Figure \ref{fig:frontierstats} that, on Frontier, both the model uncertainty in US and the committee disagreement in QC lead to the selection of slower configurations at the beginning. After the initial selections, however, QC moves toward batches with smaller mean runtimes than US. This indicates that the strongest disagreement over slower configurations is reduced once those batches are labeled, after which the committee disagreement shifts toward faster configurations. US remains in a range, around 150--300 s, which implies that model uncertainty continues to occur across both smaller and larger runtime configurations. RS continues to select batches with average runtime.

Figure \ref{fig:perlmutterstats} shows that compared with Aurora and Frontier, both QC and US on Perlmutter explore batches with smaller runtimes in the early stage. However, QC and US follow the same behavior of selecting more expensive configurations before transitioning toward batches with smaller average runtimes in the later iterations.

\subsubsection*{Dispersion}
\begin{figure*}[t]
  \centering
    \includegraphics[scale=0.4]
    {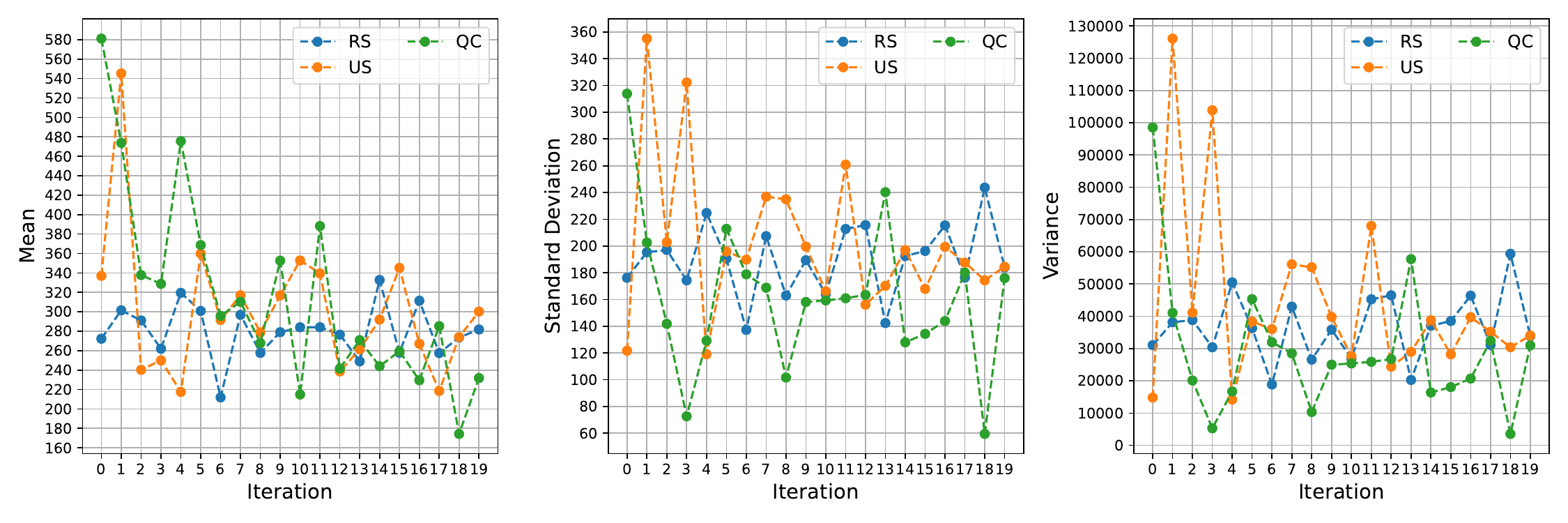}
    \vspace{-20pt}
    \caption{Statistics of the active learning results --- Aurora}
    \label{fig:aurorastats}
\end{figure*}

From Figure \ref{fig:aurorastats}, QC begins with a large standard deviation and variance on Aurora, consistent with its large initial mean runtime. This indicates that the configurations committee first disagrees on the most are not only slower on average but also varied in runtime. After that, both the standard deviation and variance of QC decrease, which, combined with the low mean, illustrate that the model is now learning batches with similar configurations with small runtimes. US, however, continues to show a larger standard deviation and variance across several iterations compared to QC. This suggests that model uncertainty on Aurora is caused by a wider range of runtime configurations, while QC becomes more homogeneous after its initial selections. RS consistently selects randomly, so it is close to the mean across iterations.

\begin{figure*}[t]
  \centering
  \includegraphics[scale=0.4]{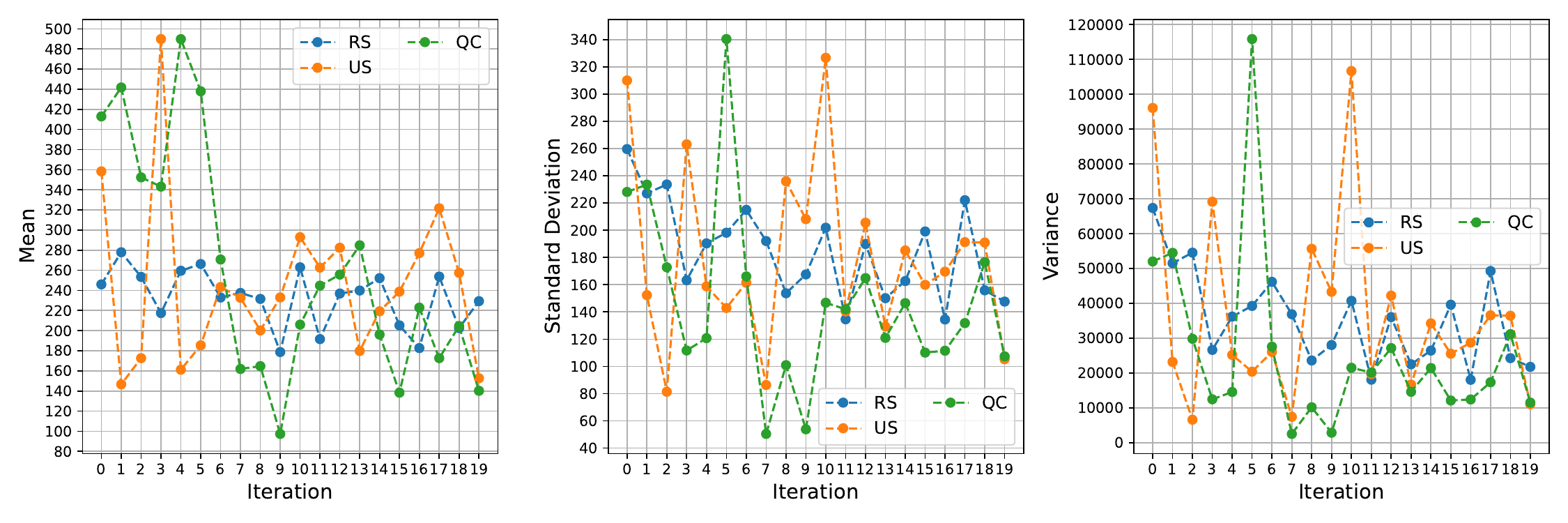}
  \vspace{-20pt}
    \caption{Statistics of the active learning results --- Frontier}
    \label{fig:frontierstats}
\end{figure*}

Figure \ref{fig:frontierstats} shows that, on Frontier, QC has larger runtime variation in the early iterations, but after the initial selections, its standard deviation and variance decrease. This, in combination with the mean-runtime result, shows that QC first selects slower and more varied batches. After those configurations are learned, QC selects batches that are faster on average and more similar in runtime. Similar to Aurora, US maintains larger standard deviation and variance across more iterations, which suggests that the configurations with the largest uncertainty estimates continue to include a wider range of runtime values, whereas QC selects batches with more similar runtimes after the early iterations.

\begin{figure*}[t]
  \centering
  \includegraphics[scale=0.4]{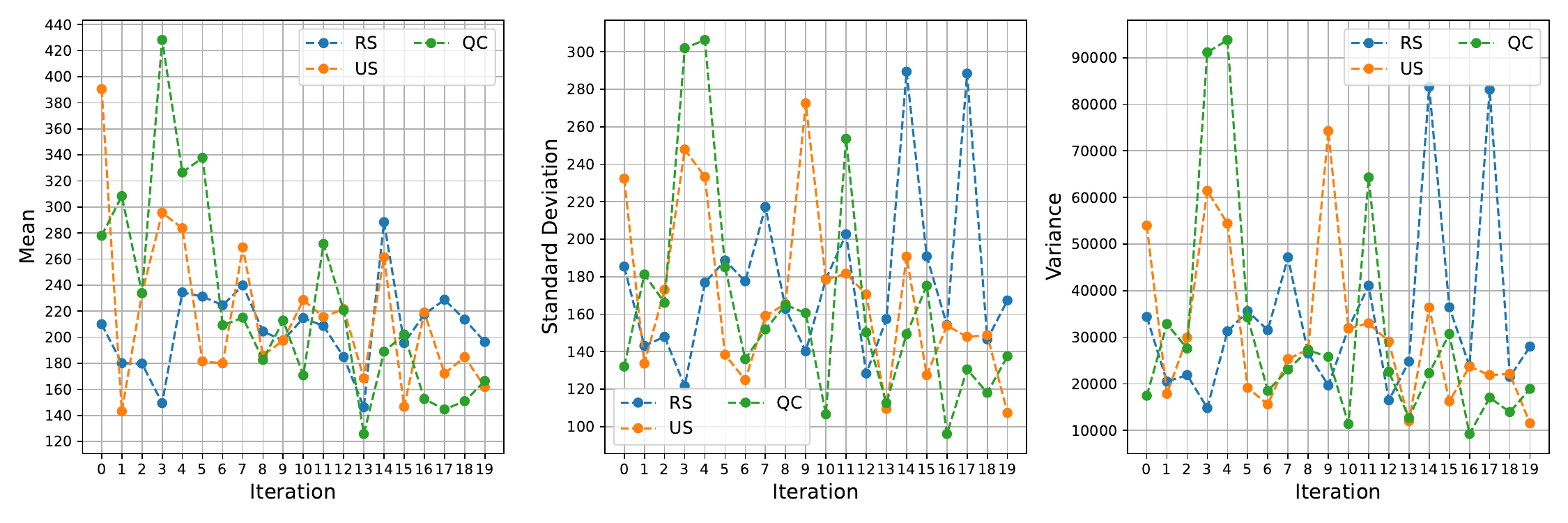}
  \vspace{-20pt}
    \caption{Statistics of the active learning results --- Perlmutter}
    \label{fig:perlmutterstats}
\end{figure*}

Figure \ref{fig:perlmutterstats} shows that QC starts with smaller standard deviation and variance on Perlmutter than on Aurora and Frontier. Later increases in QC dispersion show that the configurations with the largest disagreement scores are not limited to runs with similar execution times. Instead, the selected batches include both shorter and longer runs, so QC continues to learn information from different parts of the runtime distribution. US also shows increases in standard deviation and variance at several iterations. For these batches, the model is uncertain about configurations with a wide range of runtimes. Interestingly, RS also shows larger later increases in standard deviation and variance on Perlmutter compared with Aurora and Frontier, implying a higher runtime variability present in the Perlmutter batches sampled at those iterations.

\subsection{Timing analysis of generated samples}
In this section, we will study the behavior of the generative learning methods as the amount of real data \(n\) grows, and as we vary the number of generated samples \(m \in \{n,2n,3n\}\). In Figures~\ref{fig:auroraGCstat}--\ref{fig:perlmutterctganstat}, we plot the mean, standard deviation, and variance of synthetic runtimes in seconds for each 
\((n,m)\). The mean is the average runtime (seconds) for a given generated batch \((n,m)\). A higher mean indicates that the generated experiment batches are slower for that configuration, whereas a lower mean indicates faster batches. The standard deviation measures how much the generated runtimes deviate from their mean. Variance is average of the squared differences from the mean. Larger values of standard deviation and variance mean the generator is producing heterogeneous experiments that combine very fast and very slow runs while smaller values mean more uniform batches clustered around the mean. We plot standard deviation and variance across iterations to observe whether the generated batches contain a wide range of experiments or the algorithm is generating within a narrow range.

\subsubsection*{Mean}
\begin{figure*}[t]
  \centering
  \includegraphics[scale=0.4]{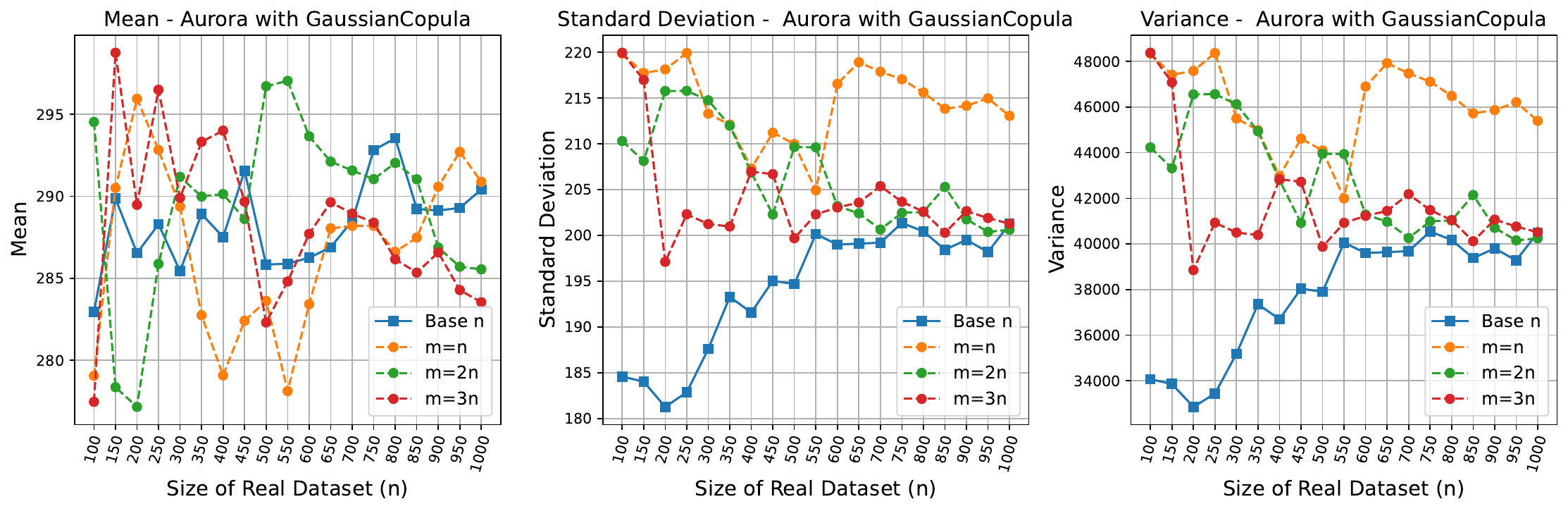}
    \caption{Statistics of the generative learning results --- Aurora with GaussianCopula}
    \label{fig:auroraGCstat}
\end{figure*}

\begin{figure*}[t]
  \centering
  \includegraphics[scale=0.4]{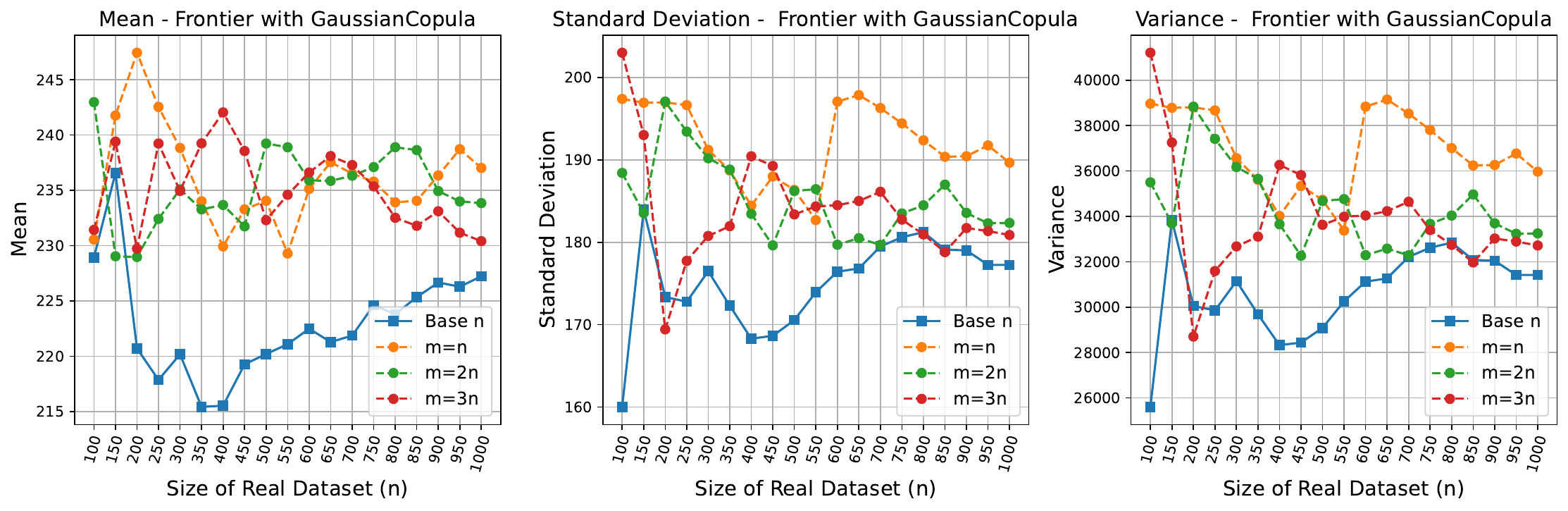}
    \caption{Statistics of the generative learning results --- Frontier with GaussianCopula}
    \label{fig:frontierGCstat}
\end{figure*}

\begin{figure*}[t]
  \centering
  \includegraphics[scale=0.4]{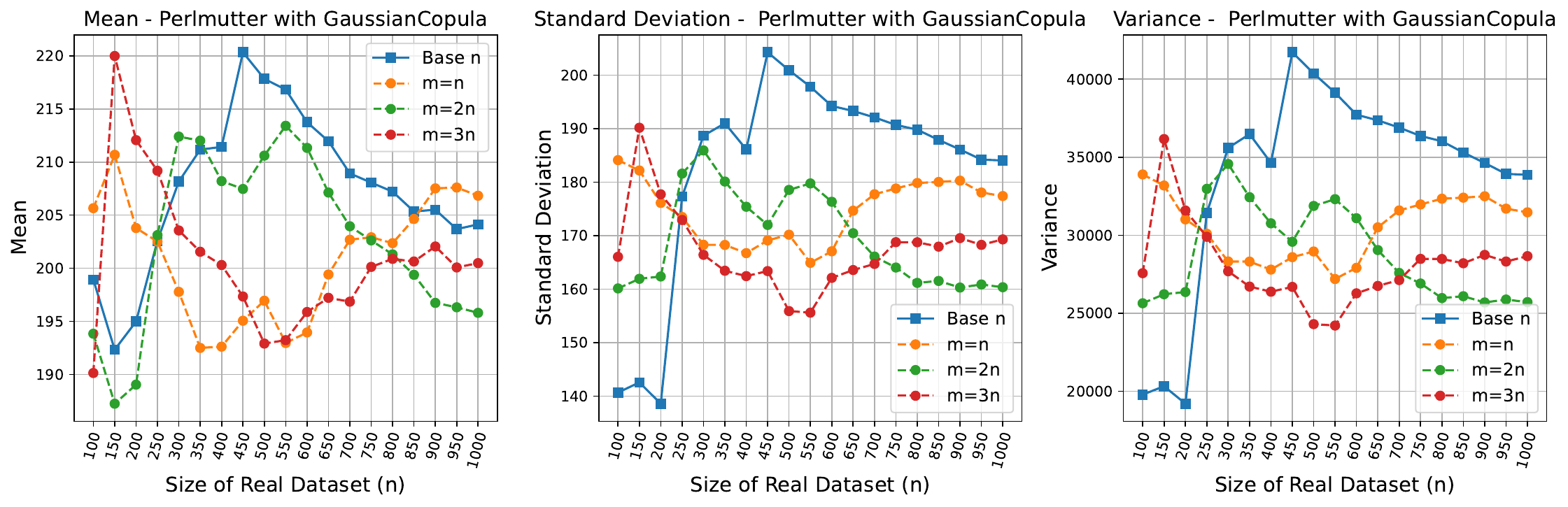}
    \caption{Statistics of the generative learning results --- Perlmutter with GaussianCopula}
    \label{fig:perlmutterGCstat}
\end{figure*}

Figures \ref{fig:auroraGCstat}, \ref{fig:frontierGCstat}, and \ref{fig:perlmutterGCstat} show the runtime statistics of the synthetic data generated by Gaussian Copula for all three machines. We can see that the mean runtime of the synthetic data is within a small range across the three sampling levels of 1, 2, and 3$\times$ of the number of the known data points. For Aurora, the mean remains within 277--297s, for Frontier, it remains within 229--247.5s, and for Perlmutter, it is 187--220s. The range becomes even smaller after the number of known data points reaches about 600. Compared with the Base n curve, Gaussian Copula generally preserves the mean runtime more closely than CTGAN. For Perlmutter, m=2n follows the Base n curve more closely than the other Gaussian Copula sampling levels for most values of n.

\begin{figure*}[t]
  \centering
  \includegraphics[scale=0.4]{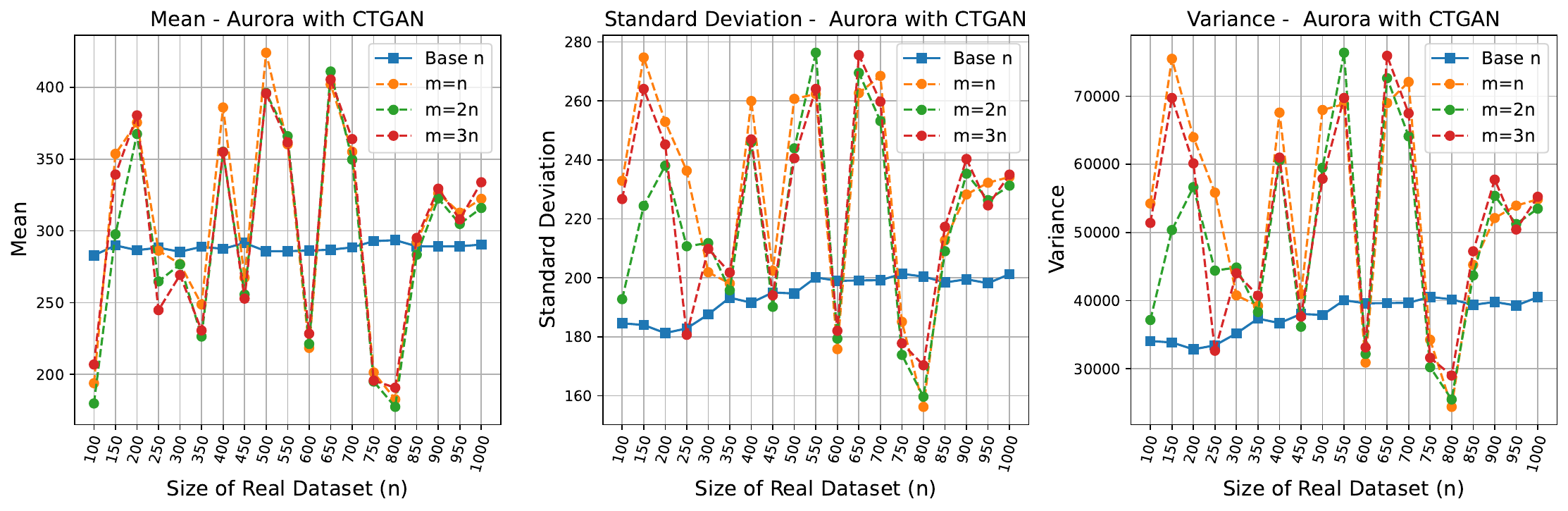}
    \caption{Statistics of the generative learning results --- Aurora with CTGAN}
    \label{fig:auroractganstat}
\end{figure*}

\begin{figure*}[t]
  \centering
  \includegraphics[scale=0.4]{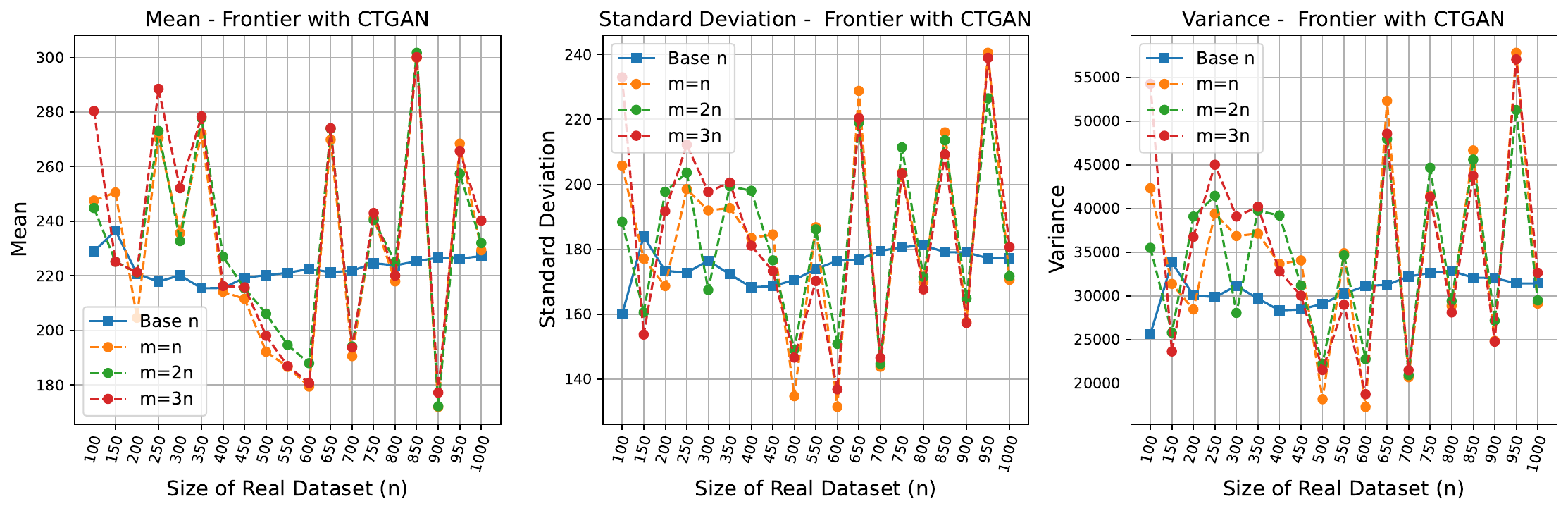}
    \caption{Statistics of the generative learning results --- Frontier with CTGAN}
    \label{fig:frontierctganstat}
\end{figure*}

\begin{figure*}[t]
  \centering
  \includegraphics[scale=0.4]{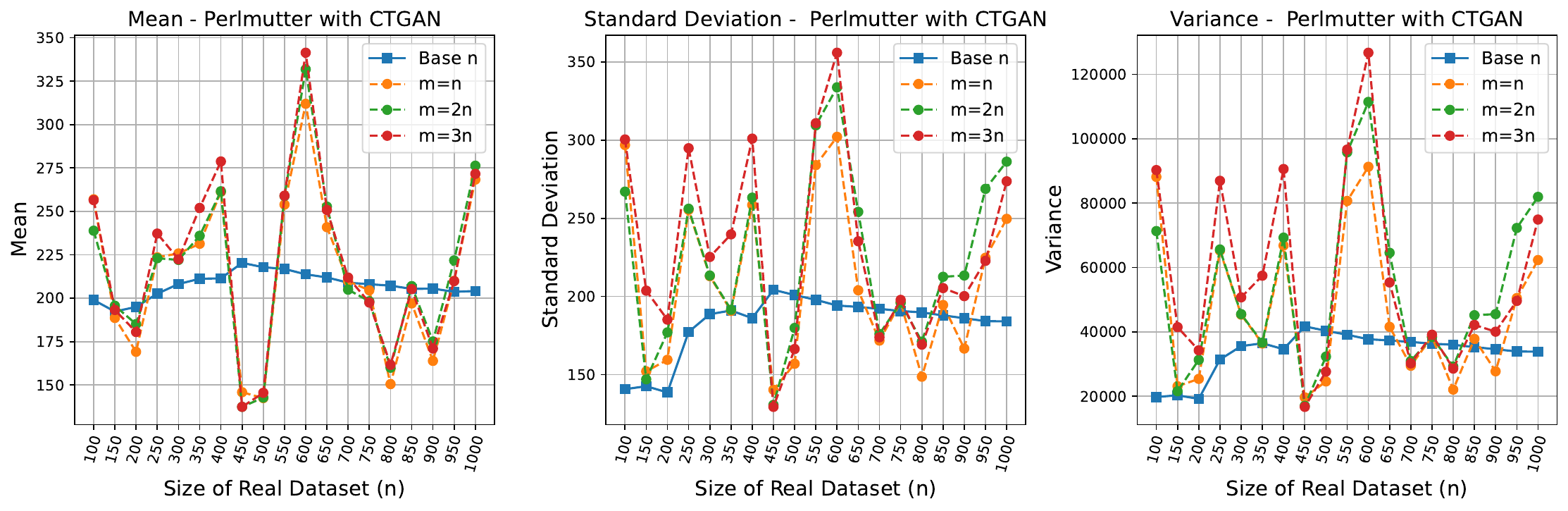}
    \caption{Statistics of the generative learning results --- Perlmutter with CTGAN}
    \label{fig:perlmutterctganstat}
\end{figure*}

The mean runtime of the synthetic data generated by CTGAN for Aurora, Frontier, and Perlmutter is shown in Figures \ref{fig:auroractganstat}, \ref{fig:frontierctganstat}, and \ref{fig:perlmutterctganstat}. Compared with Gaussian Copula, CTGAN produces synthetic datasets with larger runtime variation. The CTGAN curves also deviate more from the Base n curve, especially on Aurora and Perlmutter, where the synthetic mean runtimes move above and below the baseline by larger amounts. Unlike Gaussian Copula, increasing the number of synthetic samples for a particular n does not introduce a meaningful change in the runtime distribution, since the m=n, m=2n, and m=3n curves follow similar patterns.

\subsubsection*{Dispersion}
After 600 data points, generating 2 and 3$\times$ synthetic data samples relative to the number of known data points produces less variable data compared to 1$\times$ data for Gaussian Copula. Compared to the Base n curve, the Gaussian Copula dispersion is close to the baseline on Aurora, slightly higher than the baseline on Frontier, and lower than the baseline on Perlmutter after the smaller known data sizes.

CTGAN shows high variability in the synthetic samples generated across different known data sizes. Compared to the baseline, CTGAN shows larger deviations in standard deviation and variance than Gaussian Copula.




\section{Conclusions}
\label{conclusion}

In this work, we presented a comprehensive ML framework for predicting optimal runtime parameters of large‑scale CCSD computations on premier supercomputing platforms. By evaluating a broad set of regression models, we identified Gradient Boosting as the most accurate and robust predictor across Aurora, Frontier, and Perlmutter, achieving high accuracy when sufficient data are available. To address data scarcity where running new experiments is expensive, we utilized active learning and generative learning. Our active‑learning methods, particularly US, achieved strong accuracy with only 20--35\% of the original data, substantially reducing the number of required supercomputer runs. Generative learning, including Gaussian Copula and CTGAN, further improved performance in some settings.

For both the STQ and Budget questions, our models showed consistent ability to recover near‑optimal configurations. Moreover, cross‑platform comparisons revealed prediction performance differences among systems. Models trained with Frontier experiments outperformed those with both Aurora and Perlmutter on most common configurations.

Overall, our results show that active and generative learning strategies can greatly reduce the burden of parameter selection in large‑scale quantum chemistry simulations. Our approach enables more efficient resource usage, faster configuration tuning, and improved user decision‑making.

\section*{Acknowledgements}
Acknowledgements, funding information, and author contribution details will be provided through the journal submission system and added to the manuscript after peer review.

\section*{Data and code availability}
All input and resulting data, including code, will be made available upon request.

\section*{Ethical statement}
This article does not contain any studies with human participants or animals performed by any of the authors.

\bibliographystyle{iopart-num}
\bibliography{refs}
\end{document}